%% file: clean-arxiv.tex
\documentclass{article} 
\usepackage{iclr2023_conference,times}

\input{math_commands.tex}

\usepackage{amsmath}
\usepackage{amssymb}
\usepackage{graphicx}
\usepackage{hyperref}
\usepackage{float}
\usepackage{bbm}

\newcommand{\bs}{\boldsymbol}

\newcommand{\airloc}{AiRLoc}
\newcommand{\bth}{\bs{\theta}}
\newcommand{\pith}{\pi_{\bth}}

\newcommand{\Figure}[1]{Fig.~\ref{#1}}
\newcommand{\Table}[1]{Table~\ref{#1}}
\newcommand{\Section}[1]{\S\ref{#1}}

\begin{document}
\title{Aerial View Localization with Reinforcement Learning: \emph{Towards Emulating Search-and-Rescue}}
%
%



\author{Aleksis Pirinen$^1$, Anton Samuelsson$^2$, John Backsund$^2$ and Kalle Åström$^{2}$\\
$^1$RISE Research Institutes of Sweden\\
$^2$Centre for Mathematical Sciences, Lund University, Sweden \\
\tt\small \{aleksis.pirinen@ri.se, anton.b.samuelsson@gmail.com, \\ 
\tt\small j.backsund@gmail.com, karl.astrom@math.lth.se\}}

%

\newcommand{\fix}{\marginpar{FIX}}
\newcommand{\new}{\marginpar{NEW}}

\iclrfinalcopy 

\maketitle              
\begin{abstract}
Climate-induced disasters are and will continue to be on the rise, and thus search-and-rescue (SAR) operations, where the task is to localize and assist one or several people who are missing, become increasingly relevant. In many cases the rough location may be known and a UAV can be deployed to explore a given, confined area to precisely localize the missing people. Due to time and battery constraints it is often critical that localization is performed as efficiently as possible. In this work we approach this type of problem by abstracting it as an \emph{aerial view goal localization} task in a framework that emulates a SAR-like setup without requiring access to actual UAVs. In this framework, an agent operates on top of an aerial image (proxy for a search area) and is tasked with localizing a goal that is described in terms of visual cues. To further mimic the situation on an actual UAV, the agent is not able to observe the search area in its entirety, not even at low resolution, and thus it has to operate solely based on partial glimpses when navigating towards the goal. To tackle this task, we propose \emph{\airloc{}}, a reinforcement learning (RL)-based model that decouples exploration (searching for distant goals) and exploitation (localizing nearby goals). Extensive evaluations show that \airloc{} outperforms heuristic search methods as well as alternative learnable approaches, and that it generalizes across datasets, e.g.~to disaster-hit areas without seeing a single disaster scenario during training. We also conduct a proof-of-concept study which indicates that the learnable methods outperform humans on average. Code and models have been made publicly available at \url{https://github.com/aleksispi/airloc}.
\end{abstract}
%
%
%
\section{Introduction}
\begin{figure}[t]
    \centering
    \includegraphics[width=0.99\textwidth]{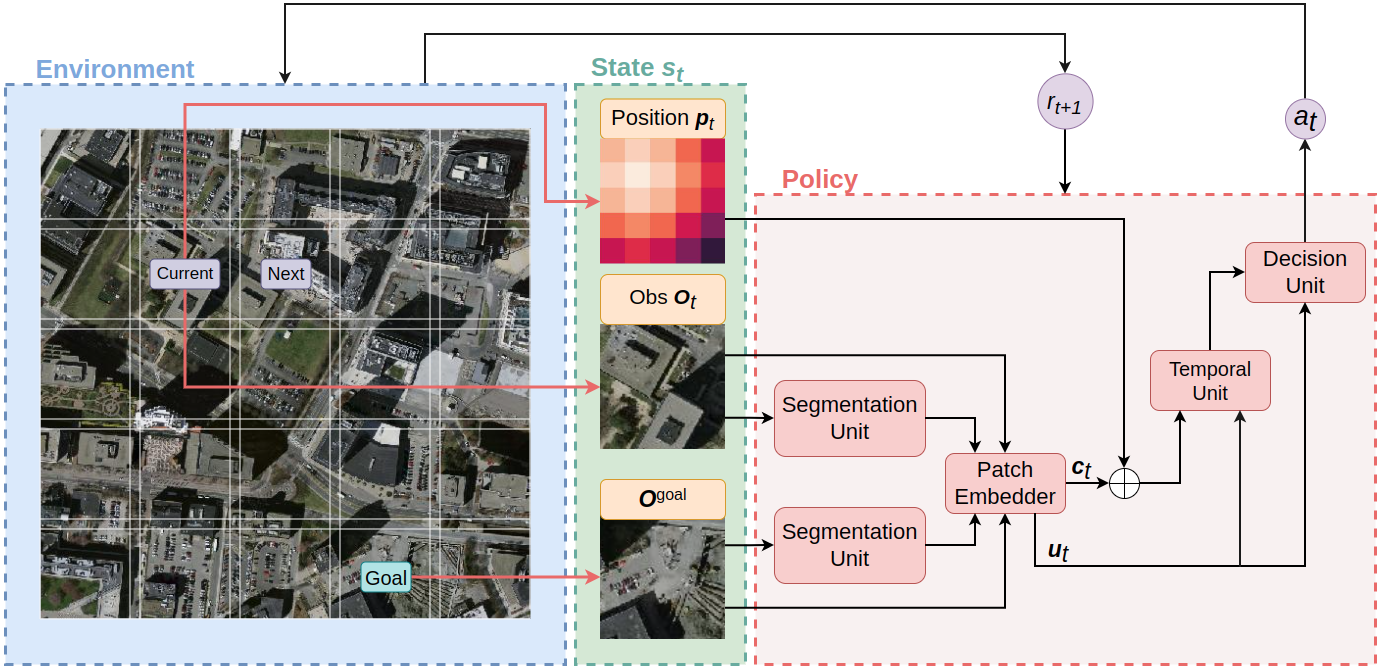}
    \vspace{-5pt}
    \caption{Overview of \emph{AiRLoc}, our RL-based agent for aerial view goal localization. The state $s_t$ consists of the agent's current position $\bs{p}_t$, its currently observed patch $\bs{O}_t$, and the goal patch $\bs{O}^\text{goal}$. First, segmentation masks for $\bs{O}_t$ and $\bs{O}^\text{goal}$ are computed, and $\bs{O}_t$, $\bs{O}^\text{goal}$ and their segmentations are then fed through a patch embedder to generate a common representation $\bs{c}_t$. The positional encoding $\bs{p}_t$ is then added to $\bs{c}_t$, and the sum, together with an exploitation prior $\bs{u}_t$ (see \Section{sec:airloc}), are subsequently processed by an LSTM, whose output is fed to a decision unit. The decision unit also receives $\bs{u}_t$ and outputs an action probability distribution $\pi(\cdot | s_t)$. A movement $a_t$ is then sampled from $\pi(\cdot | s_t)$, which results in the next state $s_{t+1}$ and reward $r_{t+1}$ (a reward is provided during training only). The process is repeated, either until the agent reaches the goal, or until a maximum number of steps $T$ have been taken. Note that \airloc{} never observes the full search area, not even at a low resolution.}
    \label{fig:DetailedPolicy}
    \vspace{-9pt}
\end{figure}

Recent technological developments of unmanned aerial vehicles (UAVs) and satellites have resulted in an enormous increase in the amount of aerial view landscape and urban data that is available to the public \citep{landcover,masa_dataset,dubai,kuzin2021disaster,xiong2022earthnets,schmitt2022eod,xia2022openearthmap}. An important application area of UAVs is within search-and-rescue (SAR) operations, where the task is to localize and assist one or several people who are missing, for example after a natural disaster. It may often be the case that the people in need are known to be within a confined area, such as within a specific neighborhood or city block. In such a scenario, a UAV can be used to explore the area from an aerial perspective to precisely localize and subsequently assist the missing people. Obviously, controlling the UAV in an informed and intelligent manner, rather than exhaustively scanning the whole area, could significantly improve the likelihood of succeeding with the operation.
\\ \\
In this paper, we propose a novel setup and task formulation that allows for controllable and reproducible development of and experimentation with systems for UAV-based SAR operations.\footnote{Also relevant for many types of environmental monitoring applications, e.g.~in forestry management.} More specifically, we abstract the problem within a framework that emulates a SAR-like setup without requiring access to actual UAVs.
In this framework, an agent operates on top of an aerial image (proxy for a specific search area) and is tasked with localizing a goal for which coordinates are not available, but where some visual cues of the goal are provided. For our task, which we denote \emph{aerial view goal localization}, we assume that the visual cues are given in terms of a top-view observation of the goal within the search area (see \Figure{fig:DetailedPolicy}). 
This provides a streamlined proxy setup, but note that in a real SAR operation such cues could instead be provided e.g.~by the missing people, assuming they have been able to send information about their surroundings (e.g.~ground-level images). The active localization methodologies we propose can easily be extended to allow for more flexible goal specifications, for example by integrating an off-the-shelf geo-localization module.
\\ \\
There are many cases where GPS coordinates of the goal location are not available, or where such information is not reliable (e.g.~because global satellite navigation systems are susceptible to radio frequency interruptions and fake signals). Hence there is a need for robust aerial localization systems that do not rely on global positional information, but that can operate reliably based on visual information alone. Moreover, to further mimic the situation on an actual UAV, it is assumed in our task that only a partial glimpse of the search area can be observed at the same time. In many cases, a UAV could elevate to a higher altitude to get a generic (lower-resolution) sense of the whole search area, but there are also conditions which makes this impractical, e.g.~if the battery of the UAV is running low. Adverse weather conditions could also make it risky or impossible to operate at a high altitude.
\\ \\
To tackle our suggested aerial view goal localization task, we propose \emph{\airloc{}}, a reinforcement learning (RL)-based model that decouples exploration (searching for distant goals) and exploitation (localizing nearby goals) -- see \Figure{fig:DetailedPolicy}. Extensive experimental results show that \airloc{} outperforms heuristic search methods and alternative learnable approaches. The results also show that \airloc{} generalizes across datasets, e.g.~to disaster-hit areas without seeing a single disaster scenario during its training phase. We also conduct a proof-of-concept study which indicates that this task is difficult even for humans.

\section{Related Work}
Several prior works have proposed methods for autonomous control a UAVs \citep{stache2022adaptive,area1,area2,astar,sadat2015fractal,urbanmapping,popovic2020informative}. Many of these works (e.g.~\cite{stache2022adaptive,sadat2015fractal,urbanmapping}) revolve around methodologies for efficient scanning of large areas (e.g.~agricultural landscapes) such that certain types of global-level downstream inferences -- such as determining the health status of a field of crops -- can be accurately performed based on a limited number of high-resolution observations. Aside from differing in task formulation (ours requiring precise localization of a particular goal, while the aforementioned works often revolve around global-level inference), these prior works assume access to a global lower-resolution observation of the whole area of interest, while we do not. There are also works that are closer to us in terms of task setup \citep{astar,area1,area2}. For example, \cite{astar} propose a hierarchical planning approach for a goal reaching task, where a rough plan is first proposed using A*. This rough plan is subsequently used as an initial guess by a finer-grained planner which parametrizes the initial trajectory as continuous B-splines and performs trajectory optimization. Different from us, their system assumes access to ground truth detections of moving objects and ground classifications.
\\ \\
Our work is also related but orthogonal to the increasingly studied problem of geo-localization \citep{wilson2021visual,danish,satellite,ground2sat,pramanick2022world,wang2022multiple,shi2022beyond,berton2022deep,berton2022rethinking,zhu2022transgeo,downes2022city}. Such works aim to infer relationships between two or more images from different perspectives, e.g.~predicting the satellite or drone view corresponding to a ground-level image. Most such methods perform this task by an exhaustive comparison within a large image set, and are thus very different to our setup which instead revolves around minimizing the amount of observations when performing localization. However, our proposed methodologies could further benefit from incorporating geo-localization methods. For example, if the goal location is specified from a ground-level perspective, which may be more realistic in practice, geo-localization methods can be used to match the top-view images observed by our proposed method during goal localization. 
\\ \\
From a pure task formulation perspective, and setting aside the application areas, our setup may be most closely related to embodied image goal navigation \citep{anderson2018evaluation,zhu2017target,memorymodule}. In this framework, an agent is tasked to navigate in a first-person perspective within a 3d environment towards a goal location which is specified as an image within the environment. On the one hand, the embodied setting may sometimes be more challenging than our setup, since the exploration trajectories are typically longer (as the agent moves a significantly smaller extent per action) and because exploration is performed among obstacles (e.g.~walls and furniture). On the other hand, embodied first person agents may often observe the goal from far away (e.g.~from the other side of a newly entered room), while our formulation is more challenging in that the goal can never be observed in any way prior to reaching it.
\\ \\
To the best of our knowledge, in addition to us relatively few prior works have considered inference based solely on partial glimpses of an underlying image \citep{rangrej2021probabilistic,rangrej2022consistency}. In contrast, most earlier RL-based methods that have been proposed for computer vision tasks -- e.g.~for object detection \citep{caicedo2015active,gao2018dynamic,pirinen2018deep} and aerial view processing \citep{uzkent2020learning,ayush2020efficient} -- assume access to at least a low-resolution version of the entire scene or image being processed. Even the seminal work by \cite{mnih2014recurrent} uses lower-resolution full image input in addition to high-resolution partial glimpses during its sequential processing, even though in principle it may be possible to re-design the system to operate based on high-resolution glimpses alone.
\section{Aerial View Goal Localization}\label{sec:task-description-start}
In this section we first explain in detail our proposed aerial view goal localization task and framework (\Section{sec:task-description}). Then, in \Section{sec:airloc}, we explain \airloc{}, our reinforcement learning (RL)-based approach for tackling this task. See \Figure{fig:DetailedPolicy} for an overview. Finally, \Section{sec:baselines} describes the baseline methods we have developed and that we evaluate and compare with \airloc{} in \Section{sec:experiments}.
\subsection{Task Description}\label{sec:task-description}
The task is executed by an agent within a \emph{search area}, which is discretized as an $M \times N$ grid that is layered on top of a given aerial image (with a small distance between each grid cell, to avoid overfitting models to edge artefacts). Every grid cell within the search area corresponds to a valid position $\bs{p}_t$ of the agent, and the agent can only directly observe the image content $\bs{O}_t$ of its current cell. In each episode, one of the grid cells corresponds to the goal that the agent should localize. The image content of the goal cell is denoted $\bs{O}^\text{goal}$ and its position is denoted $\bs{p}^\text{goal}$. Note that the goal position $\bs{p}^\text{goal}$ is \emph{never} observed by the agent; it is only used to determine if the agent is successful. The task is considered successfully completed as soon as the agent's current position $\bs{p}_t$ and the goal position $\bs{p}^\text{goal}$ coincide,\footnote{A reasonable next step would be to require that an agent has to declare when it has reached its goal.} i.e.~when $\bs{p}_t = \bs{p}^\text{goal}$.
\\ \\
In each episode, the agent's start location $\bs{p}_0$ and the goal location $\bs{p}^\text{goal}$ are sampled at uniform random within the search area ($\bs{p}_0 \neq \bs{p}^\text{goal}$). The agent then moves around until it either reaches the goal ($\bs{p}_t = \bs{p}^\text{goal})$, or a maximum number of steps $T$ have been taken. This limit $T$ is included to represent time and resource constraints. In our task formulation, an agent has eight possible actions, which correspond to moving to any of its eight adjacent locations (grid cells). An agent may in general move outside the search area, and if so, the agent receives an entirely black observation. There is never any advantage to moving outside the search area, and thus it should be avoided (it is easy to avoid given $\bs{p}_t$).
\subsection{AiRLoc Model}\label{sec:airloc}
In this subsection we describe \emph{AiRLoc}, the reinforcement learning (RL)-based model we propose for tackling the aerial view goal localization task. An overview is shown in \Figure{fig:DetailedPolicy}.
\\ \\
\textbf{States, actions and rewards.}
The state $s_t$ contains the currently observed patch $\bs{O}_t$, the goal patch $\bs{O}^\text{goal}$, and an encoding $\bs{p}_t \in \mathbb{R}^{256}$ of the agent's position. As described above, \airloc{} has eight possible actions $a_t$, which correspond to moving to any of its adjacent locations. During training, a negative reward is provided for each action that does not move the agent into the goal location, and a positive reward is provided when the goal is found. Specifically, after taking action $a_{t-1}$ in state $s_{t-1}$ the reward $r_t = 3 \cdot \mathbbm{1}\left(\bs{p}_t = \bs{p}^\text{goal}\right) - 1$ is provided, where $\mathbbm{1}$ is the indicator function.
\\ \\
\textbf{Policy overview:} In each step, the state $s_t$ is processed by four modules to generate the current action distribution $\pith(\ast | s_t)$, where $\bth$ denotes all learnable parameters. First, $\bs{O}_t$ and $\bs{O}^\text{goal}$ are passed through a pretrained \textit{segmentation unit} (a U-net \citep{UNet}, see supplement) which predicts building segmentation masks for $\bs{O}_t$ and $\bs{O}^\text{goal}$, respectively. Second, $\bs{O}_t$ and $\bs{O}^\text{goal}$ and their segmentations are passed through a \textit{patch embedder} which yields a low-dimensional embedding $\bs{c}_t \in \mathbb{R}^{256}$ of what the agent observes and what it aims to localize. The patch embedder also outputs an exploitation prior $\bs{u}_t \in \mathbb{R}^8$ (described more below). Third, $\bs{p}_t$ is added to $\bs{c}_t$ and the result and $\bs{u}_t$ are passed to an LSTM-based \emph{temporal unit} \citep{LSTM} which integrates information over time. Finally, the LSTM output and $\bs{u}_t$ are passed to a \emph{decision unit} which yields the probability distribution $\pith(\ast | s_t)$. This decision unit first projects the LSTM's output into the action space dimensionality, then adds the exploitation prior $\bs{u}_t$, and finally generates an action distribution using softmax. Note that we use an LSTM rather than a Transformer for the temporal unit, since we want to keep the overall architecture lightweight -- the model weights occupy less than 4 MB of memory, and inference can be efficiently performed even without a GPU.
\\ \\
\textbf{Patch embedder:} The patch embedder should extract relevant information about the relationship between $\bs{O}_t$ and $\bs{O}^\text{goal}$. To achieve this, we use an architecture similar to that by \cite{doerch}, who consider a self-supervised visual representation learning task where the spatial displacement between a pair of adjacent random crops from an image should be predicted.
Note that when the start location $\bs{p}_0$ is adjacent to the goal location $\bs{p}^\text{goal}$, and when the movement budget $T=1$, our task becomes equivalent to the representation learning task introduced by \cite{doerch}.
Our patch embedder architecture consists of two parallel branches with four convolutional layers (ReLUs and max pooling are applied between layers). First, $\bs{O}_t$ and $\bs{O}^\text{goal}$, with their segmentations channel-wise concatenated, are fed separately into one branch each. To enable early information sharing between the agent's current patch and the goal patch, after two convolutional layers, the outputs of the two branches are concatenated and sent through the rest of their respective branches. The two resulting 128-dimensional embeddings are then concatenated and the result is passed through a dense layer with output $\bs{c}_t \in \mathbb{R}^{256}$.
\\ \\
Pretraining backbone vision components is common in RL setups, since it often yields a higher end performance \citep{sax2018mid,visionPretrain,wang2022vrl3,xiao2022masked,yadav2022offline}.
We therefore pretrain the patch embedder in the same self-supervised fashion as \cite{doerch}. During pretraining, another dense layer (with input $\bs{c}_t$) is attached to produce an 8-dimensional output $\bs{u}_t$ which is fed to a softmax function. The eight outputs correspond to the possible locations of $\bs{O}^\text{goal}$ relative to $\bs{O}_t$, assuming these are adjacent. When using the patch embedder within \airloc{}, we take advantage of both $\bs{c}_t$ and $\bs{u}_t$, cf.~\Figure{fig:DetailedPolicy}. Note that $\bs{u}_t$ can be interpreted as an \emph{exploitation prior}, as it is specifically tuned towards localizing ('exploiting') adjacent goals. Thus, feeding $\bs{u}_t$ to the temporal unit as well as directly to the decision unit allows \airloc{} to learn when to explore and when to exploit (without $\bs{u}_t$, the same policy must be able to both localize adjacent goals \emph{and} explore far-away goals). The choice of using both $\bs{c}_t$ and $\bs{u}_t$ is empirically justified in \Section{sec:ablations}.
\\ \\
\textbf{Positional encoding:} Positional information is represented similarly to Transformers \citep{transformer}; see details in the supplement. Note that \airloc{} never receives global positional information, i.e.~it is always relative to a given search area. Such information may be available during SAR within a confined area, where a UAV can keep track of its location relative to the borders of this area. Let $(x,y)$ denote the agent's coordinates within the $M \times N$-sized search area (thus $x \in \{0,\dots,M-1\}$, $y \in \{0,\dots,N-1\}$). Then the $i$:th element $p_t^i$ of the positional encoding vector $\bs{p}_t \in \mathbb{R}^d$ (with $d$ even; for us $d=256$) is given by:
\begin{equation}\label{eq:pos-enc}
    p_t^i =  \begin{cases}
    \cos{ (x / 100^{2(i-1)/(d/2)})} & \text{if } i \in \{1,\dots,d/2\} \text{ and } i \text{ is odd}\\
    \sin{ (x / 100^{2i/(d/2)})}  & \text{if } i \in \{1,\dots,d/2\} \text{ and } i \text{ is even}\\
    \cos{ (y / 100^{2(i-1)/(d/2)})} & \text{if } i \in \{d/2+1,\dots,d\} \text{ and } i \text{ is odd}\\
    \sin{ (y / 100^{2i/(d/2)})}  & \text{if } i \in \{d/2+1,\dots,d\} \text{ and } i \text{ is even}\\
    \end{cases}
\end{equation}
\textbf{Policy training.} To learn the parameters of \airloc{}, we first pretrain the patch embedder in a self-supervised fashion (without RL) as described above. We then freeze the patch embedder weights and train the rest of \airloc{} using REINFORCE \citep{REINFORCE}. We employ within-batch reward normalization based on distance left to the goal, i.e.~rewards associated with states of equal distance to the goal are grouped and normalized to zero mean and unit variance. We use a pretrained segmentation unit (one can simply use an off-the-shelf aerial view segmentation model) and it is not refined during policy training -- see the supplementary material for details.

\subsection{Baselines}\label{sec:baselines}
In \Section{sec:experiments} we compare \airloc{} with the following baselines:
\begin{itemize}
    \item \textbf{\emph{Priv random}} selects actions randomly, with two exceptions: i) it cannot move outside the search area; ii) it avoids previous locations.
    \item \textbf{\emph{Local}} selects actions by repeatedly calling the pretrained patch embedder (which assumes the goal is adjacent to the current location).
    \item \textbf{\emph{Priv local}} is the same as \emph{Local} but with the privileged movement restrictions of \emph{Priv random}.
    \item \textbf{\emph{Human}} represents the average human performance from a proof-of-concept evaluation with 19 subjects (see details in the supplementary material).
\end{itemize}

\section{Experiments}\label{sec:experiments}
In this section we extensively evaluate and compare \airloc{} and the various baselines described in \Section{sec:airloc} and \Section{sec:baselines}, respectively. First we however describe what datasets and evaluation metrics we use, explain different variants of \airloc{}, and provide some further implementation details.
\\ \\
\textbf{Datasets.} We mainly use \emph{Massachusetts Buildings} (\emph{Masa}) by \cite{masa_dataset} for development and evaluation (70\% for training; 15\% each for validation and testing). The data contains images of Boston and the surrounding suburban and forested areas. It depicts houses, roads and other clearly identifiable man-made structures, but also woods and less developed regions. The data also includes segmentation masks for buildings, which are used to separately train the segmentation unit (cf.~\Figure{fig:DetailedPolicy}) that is used by most of the learnable models in the results below. Models are also evaluated on the \emph{Dubai} dataset \citep{dubai}, which also depicts urban regions, although the surrounding areas are instead dry deserts. This dataset is hence used to assess the generalization of the various methods. Finally, we also train and evaluate on the \emph{xBD} dataset by \cite{gupta2019creating}, which contains satellite images from various regions both before (\emph{xBD-pre}) and after (\emph{xBD-disaster}) various natural distastes, e.g.~wildfires and floods. In this case the models are trained on non-disaster-hit data from \emph{xBD-pre} and evaluated on \emph{xBD-disaster}, where we also ensure that the training data depicts other geographical areas than those in \emph{xBD-disaster}. More details are found in the supplementary material.
\\ \\
\textbf{Evaluation metrics.} For performance evaluation we use the following five metrics.
\textbf{\textit{Success}} is the percentage of episodes where the goal is reached.
\textbf{\emph{Steps}} is the average number of actions taken per episode (for failure episodes this is set to the movement budget $T$).
\textbf{\emph{Step ratio}} measures the average ratio between the taken number of steps and the minimum number of steps required (lower is better). It is only computed for successful trajectories.
\textbf{\emph{Residual distance}} measures the average distance between the final location relative to the goal location in unsuccessful episodes (lower is better).
Finally, \textbf{\emph{Runtime}} is the average runtime per episode.
\\ \\
\textbf{\airloc{} variants.} We also train and evaluate several ablated variants of \airloc{}.
\textbf{\emph{No sem seg}} omits the segmentation unit and uses only RGB patches in the patch embedder (which is instead pretrained with RGB-only inputs). \textbf{\emph{No residual}} omits $\bs{u}_t$ in the decision unit, but not in the temporal unit, cf.~\Figure{fig:DetailedPolicy}. Finally, \textbf{\emph{no prior}} entirely discards the prior $\bs{u}_t$ in the architecture.
\\ \\
\textbf{Implementation details.} All methods are implemented in, trained and evaluated using PyTorch. Training \airloc{}\footnote{Details about the patch embedder and segmentation network training are found in the supplement.} takes 30h on a Titan V100 GPU. To learn the parameters of the policy networks, we use REINFORCE \citep{REINFORCE} with Adam \citep{kingma2014adam}, batch size 64, search area size $M \times N = 5 \times 5$, movement budget $T=10$, learning rate $10^{-4}$, and discount $\gamma=0.9$. The grid cells of the search areas are of size $48 \times 48 \times 3$, with 4 pixels between each other to avoid overfitting models to edge artefacts (each cell corresponds to roughly $100 \times 100$ meters). Each model is trained until convergence on the validation set (typically happens within 50k batches). We apply left-right and top-down flipping of images (search areas) as data augmentation. The \airloc{} variants are trained with five random network initializations each, and the results for the median-performing models on the validation set are reported below. \airloc{} is not seed sensitive, as shown in \Section{sec:seed-analysis}. Unless otherwise specified, all models are evaluated in deterministic mode, i.e.~the most probable action is selected in each step. All models are evaluated on the exact same start configurations for fair comparisons.
\begin{table}[h]
\centering
\caption{Results on the test set of \emph{Massachusetts Buildings} (movement budget $T=10$ and $T=14$ for setups of sizes $5 \times 5$ and $7 \times 7$, respectively). For both search area sizes, the success rate of \airloc{} is higher than for the baselines. Mid-level vision capabilities (semantic segmentation) are crucial for \airloc{}'s performance.
The standard local approach performs very poorly and is significantly improved by imposing the privileged movement constraints. The time per episode is low for all methods.}\label{table:testdiscrete_uncorrupt}
\vspace{6pt}
\scalebox{1.01}{
\begin{tabular}{|c||c|c|c|c|c|}
\hline
\textbf{Agent type}                & \textbf{Success} & \textbf{Step ratio} & \textbf{Steps} & \textbf{Res. dist.}  & \textbf{Runtime} \\ \hline \hline
\textbf{\airloc{} (5x5) } & 67.6 \% & 1.45 &  6.2 & 2.4  & 120 ms \\ \hline
\textbf{Priv local (5x5)} &  64.2 \%  & 1.59 & 6.5 & 2.4 & 117 ms  \\ \hline
\textbf{Local (5x5)}  &  24.7 \% & 1.47 & 8.1 & 7.0  & 138 ms \\ \hline
\textbf{Priv random (5x5)} &  41.0 \%     &  2.56      & 8.0    & 1.6           & 48 ms  \\ \hline
\hline
\textbf{\airloc{} (7x7) } & 59.0 \%  & 1.52 & 9.4  & 3.3 & 188 ms \\ \hline
\textbf{Priv local (7x7)} & 56.3 \% & 1.72 & 9.9 & 3.4 & 178 ms \\ \hline
\textbf{Local (7x7)}  &  17.8 \% & 1.20 & 11.9 & 8.7 & 202 ms \\ \hline
\textbf{Priv random (7x7)}             &  25.2 \%     &  1.82      & 12.3  & 3.5 & 74 ms  \\ \hline
\hline
\textbf{\airloc{} (no sem seg, 5x5)}            & 61.7 \% & 1.54 & 6.7 & 2.4          & 94 ms \\ \hline
\textbf{Priv local (no sem seg, 5x5)}                 &  61.6 \%  & 1.67 & 6.8 & 2.4 & 88 ms  \\ \hline
\textbf{Local (no sem seg, 5x5)}                                   &  20.5 \%     &  1.28 & 8.4 & 6.2  & 92 ms \\ \hline
\hline
\textbf{\airloc{} (no sem seg, 7x7) }            & 52.5 \%  & 1.61 & 10.1  & 3.5           & 141 ms \\ \hline
\textbf{Priv local (no sem seg, 7x7)}             &  51.1 \% & 1.89 & 10.2 & 3.3 & 133 ms\\ \hline
\textbf{Local (no sem seg, 7x7)}                                   &  14.1 \% & 1.37 & 12.4 & 8.0 & 136 ms \\ \hline
\end{tabular}}
\vspace{-8pt}
\end{table}
\begin{figure}[t!]
    \centering
         \includegraphics[width=0.99
         \textwidth]{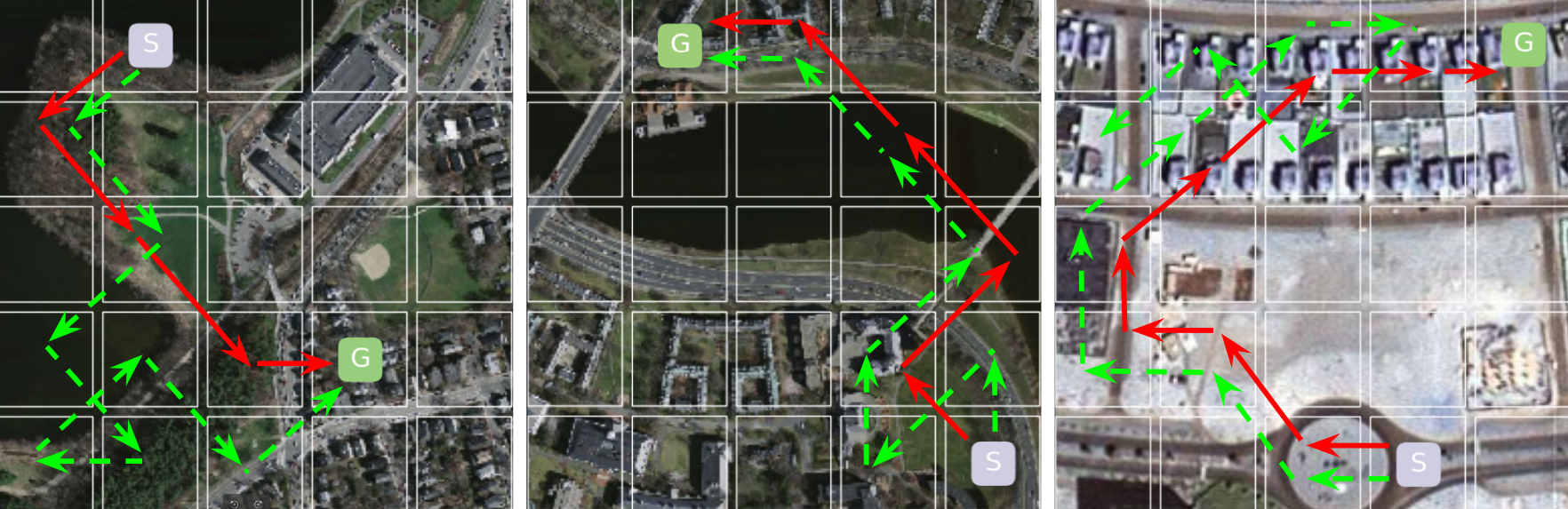}
    \vspace{-6pt}
    \caption{Examples of \airloc{} (red) and \emph{Priv local} (dashed green) on the test set of \emph{Masa} (left, middle) and \emph{Dubai} (right).
    Left: \airloc{} takes the same first two actions as \emph{Priv local} and then takes the shortest path to the goal ('G'). \emph{Priv local} also reaches the goal. Middle: \airloc{} first deviates from \emph{Priv local} and then follows the same path. \airloc{} reaches the goal faster. Right: \airloc{} follows the same path as \emph{Priv local} until it is adjacent to the goal and then moves into the goal, while \emph{Priv local} fails.}
    \label{fig:visual-example-1}
    \vspace{-6pt}
\end{figure}

\begin{table}[t!]
\centering
\caption{AiRLoc and baselines evaluated on previously unseen \emph{Dubai} data (movement budget $T=10$ and $T=14$ for setups of sizes $5 \times 5$ and $7 \times 7$, respectively). \airloc{} and the privileged local approach generalize very well to this out-of-domain data. Note that \airloc{} is the most successful method in all settings, often by a large margin.}\label{table:dubaidiscrete_uncorrupt_simple}
\vspace{6pt}
\scalebox{1}{
\begin{tabular}{|c||c|c|c|c|c|}
\hline
\textbf{Agent type}              & \textbf{Success} & \textbf{Step ratio} & \textbf{Steps} & \textbf{Res. dist.} & \textbf{Runtime} \\ \hline \hline
\textbf{\airloc{} (5x5)}            & 68.8 \% & 1.52 & 6.3 & 2.4 & 126 ms \\ \hline
\textbf{Priv local (5x5)}        & 65.6 \% & 1.59 & 6.5 & 2.4 & 113 ms \\ \hline
\textbf{Local (5x5)}                   & 23.5 \% & 1.23 & 8.2 & 6.6 & 136 ms\\ \hline
\textbf{Priv random (5x5)}       & 41.0 \% &  1.96 & 8.0 & 2.5 & 48 ms \\ \hline
\hline
\textbf{\airloc{} (7x7)}            & 57.2 \% & 1.54 & 9.7 & 3.4 & 194 ms \\ \hline
\textbf{Priv local (7x7)}        & 53.7 \% & 1.85 & 10.2 & 3.6 & 184 ms \\ \hline
\textbf{Local (7x7)}                   & 15.5 \%  & 1.25 & 12.2 & 7.9 & 207 ms\\ \hline
\textbf{Priv random (7x7)}       & 26.9 \% & 1.64 & 12.0 & 3.5 & 72 ms \\ \hline
\hline
\textbf{\airloc{} (no sem seg, 5x5)}            & 67.1 \% & 1.59 & 6.5 & 2.4 & 91 ms \\ \hline
\textbf{Priv local (no sem seg, 5x5)}        & 65.1 \% & 1.67 & 6.6 & 2.5 & 86 ms\\ \hline
\textbf{Local (no sem seg, 5x5)}                   & 23.3 \% & 1.25 & 8.2 & 6.6 & 90 ms\\ \hline
\hline 
\textbf{\airloc{} (no sem seg, 7x7)}         & 48.6 \% & 1.56 & 10.3 & 3.3 & 144 ms \\ \hline
\textbf{Priv local (no sem seg, 7x7)}        & 41.9 \% & 1.69 & 10.8 & 3.4 & 140 ms \\ \hline
\textbf{Local (no sem seg, 7x7)}             & 15.0 \% & 1.28 & 12.3 & 7.6 & 135 ms \\ \hline
\end{tabular}}
\vspace{-7pt}
\end{table}
\begin{table}[h!]
\centering
\caption{Results on scenarios depicting various natural disasters (\emph{xBD-disaster}) for models trained in two different ways. Columns 1 - 3: \airloc{} generalizes quite well from having been trained on an entirely different dataset (\emph{Masa}), which contains satellite images of non-disaster-hit urban areas, to disaster-hit areas at various other spatial locations. Columns 4 - 6: Results are improved further if models are first trained on non-disaster-hit images from the same dataset (\emph{xBD-pre}) and then evaluated at different locations depicting disaster-hit scenarios.}\label{table:results-xbd}
\vspace{6pt}
\scalebox{1.05}{
\begin{tabular}{|c||c|c|c||c|c|c|}
\hline
\textbf{Agent type}              & \textbf{Success} & \textbf{Steps} & \textbf{Runtime} & \textbf{Success} & \textbf{Steps} & \textbf{Runtime} \\ \hline \hline
\textbf{\airloc{} (5x5)}         & 66.1 \% & 6.5 & 130 ms & 72.8 \% & 6.1 & 122 ms \\ \hline
\textbf{Priv local (5x5)}        & 63.8 \% & 6.7 & 121 ms & 67.3 \% & 6.4 & 115 ms \\ \hline
\textbf{Priv random (5x5)}       & 40.8 \% & 7.9 & 48 ms & 40.8 \% & 7.9 & 48 ms \\ \hline
\hline
\textbf{\airloc{} (7x7)}         & 50.7 \% & 10.2 & 204 ms & 55.7 \% & 9.9 & 198 ms \\ \hline
\textbf{Priv local (7x7)}        & 50.5 \% & 10.2 & 184 ms & 53.6 \% & 10.0 & 180 ms \\ \hline
\textbf{Priv random (7x7)}       & 25.5 \% & 12.2 & 74 ms & 25.5 \% & 12.2 & 74 ms \\ \hline
\end{tabular}}
\vspace{-6pt}
\end{table}
\subsection{Main Results}
\begin{figure}[t]
    \centering
     \includegraphics[width=
         \textwidth]{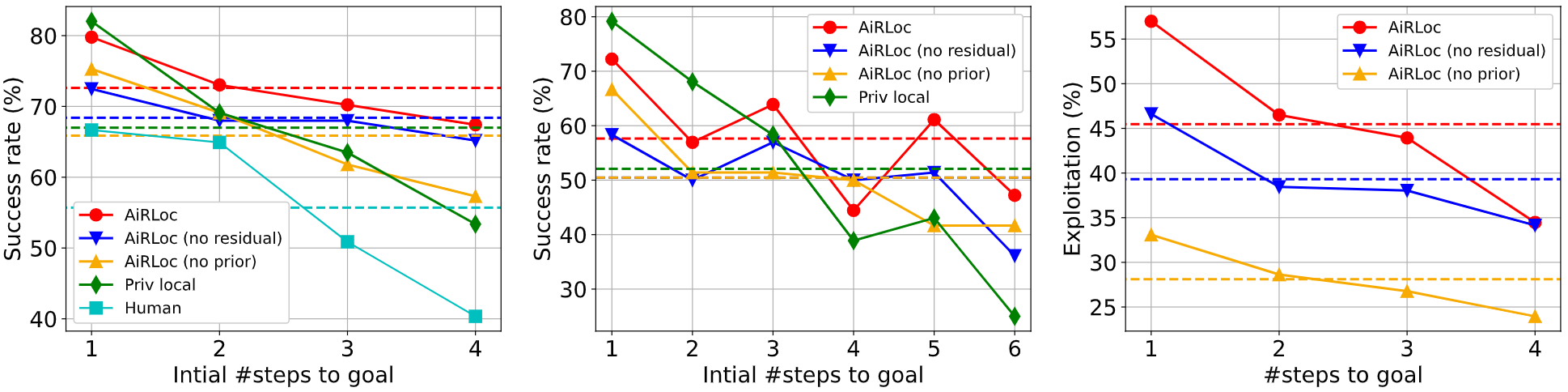}
    \vspace{-14pt}
    \caption{Left and middle: Success rate versus start-to-goal distance on the validation set of \emph{Masa} (averages are dashed). Search areas are of size $M\times N=5 \times 5$ and $T=10$ (left) or $7 \times 7$ and $14$ (middle). Left: The methods are generally more successful when the start is closer to the goal. \airloc{} and \emph{Priv local} achieve higher success rates than human operators. \airloc{} performs roughly on par with \emph{Priv local} when the goal and start are adjacent (\emph{Priv local} is trained only in this setting) and outperforms it at larger distances. \airloc{} is also more successful than its ablated variants in all settings. Middle: \airloc{} is best on average, despite having only been trained in the $5 \times 5$ setting. \emph{Priv local} is better when the start and goal are close to each other, while \airloc{} is better when they are three or more steps apart.
    Right: How frequently \airloc{} selects the same action as the exploitation prior (argmax of $\bs{u}_t$) versus goal distance. The full \airloc{} agent has the largest variability in exploitation versus exploitation depending on distance to goal.}
    \label{fig:result_diff_simple}
    \vspace{-7pt}
\end{figure}

\begin{figure}
    \centering
         \includegraphics[width=.985
         \textwidth]{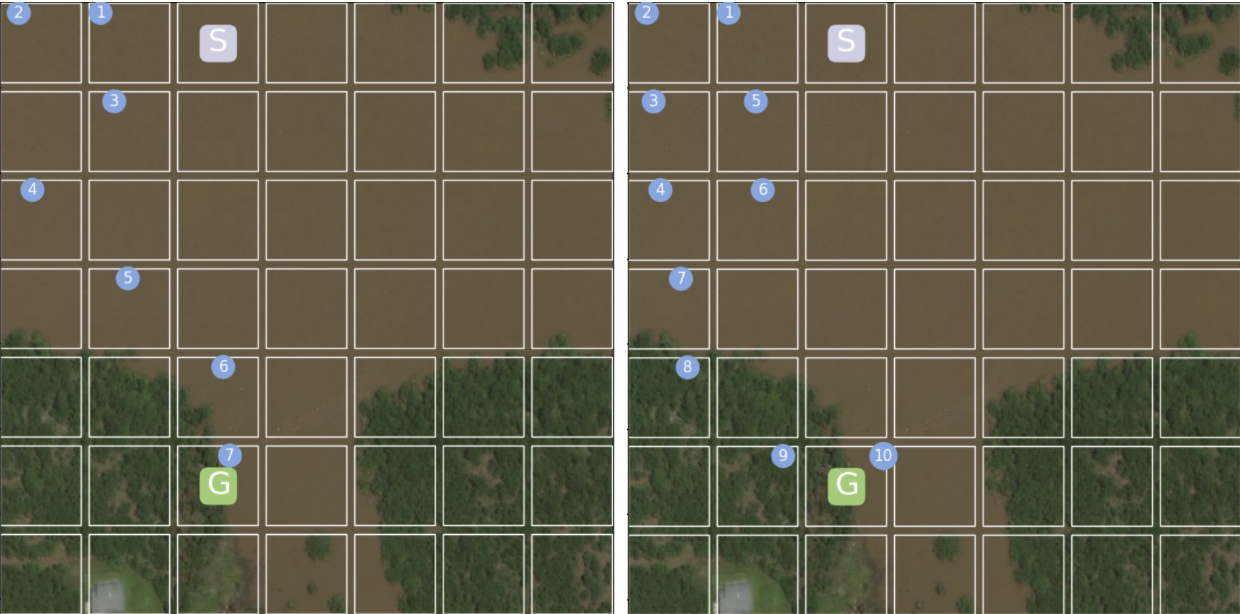}
    \vspace{-5pt}
    \caption{Successful examples of \airloc{} (left) and \emph{Priv local} (right) on a flooding scenario in \emph{xBD-disaster} ($7 \times 7$ setup, movement budget $T=14$). The start and goal locations are denoted 'S' and 'G', respectively. The numbered circles show which locations are visited and in what order. Recall that the full underlying search area is never observed in its entirety, i.e.~the agents must operate based on partial glimpses alone. Also note that \airloc{} was only trained on search areas of size $5 \times 5$ and movement budget $T=10$. \airloc{} takes the same first two steps as \emph{Priv local}, then deviates and reaches the goal in fewer steps than \emph{Priv local}.}
    \label{fig:visual-example-main-2}
    \vspace{-7pt}
\end{figure}

\begin{figure}[h!]
    \centering
         \includegraphics[width=.985
         \textwidth]{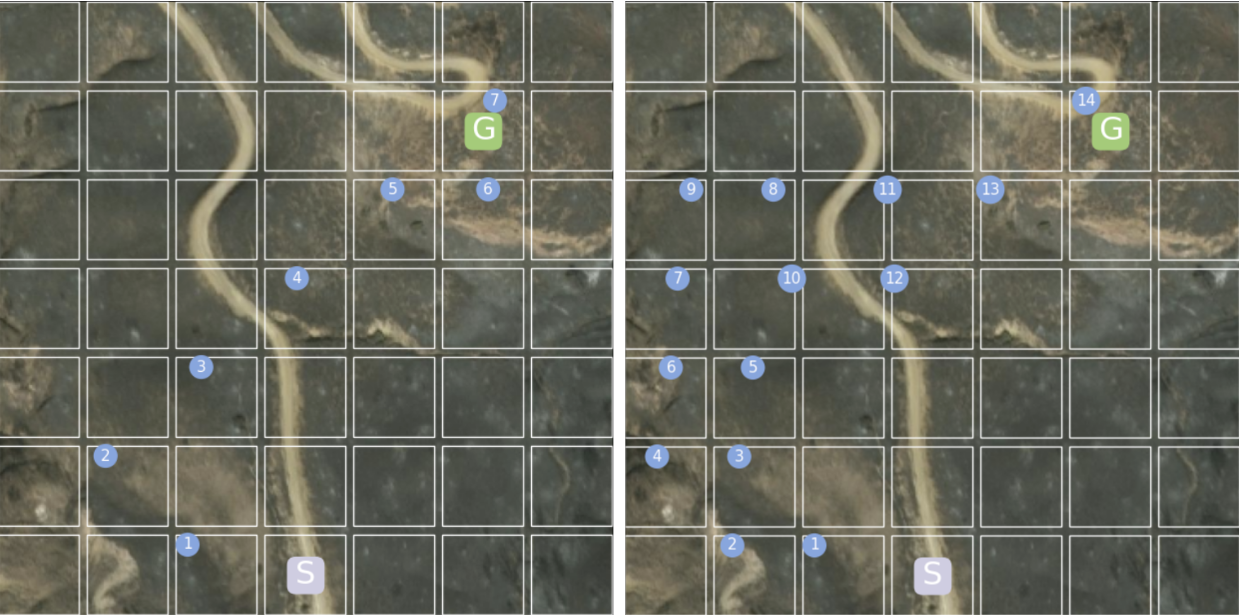}
    \vspace{-5pt}
    \caption{Successful examples of \airloc{} (left) and \emph{Priv local} (right) on a post-wildfire scenario in \emph{xBD-disaster} ($7 \times 7$ setup, movement budget $T=14$). \airloc{} takes the same first step as \emph{Priv local}, then deviates, and reaches the goal twice as fast. \emph{Priv local} precisely manages to reach the goal within its movement budget. \textbf{Please see the supplementary material for additional visualizations.}}
    \label{fig:visual-example-main-3}
\end{figure}

In \Table{table:testdiscrete_uncorrupt} we compare \airloc{} to the heuristic random and learnable local baselines on the test set of \emph{Massachusetts Buildings} (\emph{Masa}). \airloc{} obtains a higher success rate than the baselines, both in search areas of size $5 \times 5$ and $7 \times 7$ (\airloc{} is only trained in the $5 \times 5$ setting). \airloc{} and \emph{Priv local} have roughly the same runtime per trajectory, and note that all methods have runtimes that would be negligible compared to the movement overhead of an actual UAV. It is also clear that the segmentation model is crucial, which is in line with prior works that find that mid-level vision capabilities are important for high performance in RL-vision setups \citep{sax2018mid}. As seen in \Table{table:dubaidiscrete_uncorrupt_simple}, \airloc{} and the best alternative learnable approach \emph{Priv local} generalize excellently to an entirely new dataset.
\\ \\
\Table{table:results-xbd} contains results on \emph{xBD-disaster}; these results are particularly relevant from a perspective of SAR-operations in disaster-hit areas. Columns 1-3 show that \airloc{} generalizes quite well from having been trained on an entirely different dataset (\emph{Masa}), which depicts non-disaster-hit urban areas, to disaster-hit areas at various other spatial locations. Results are however improved further (columns 4-6) if models are first trained on non-disaster-hit images from the same dataset (\emph{xBD-pre}) and then evaluated at different locations that depict disaster-hit scenarios.
\\ \\
In summary, \airloc{} outperforms the baselines across all datasets and search area sizes, and localizes goals in fewer steps on average. See \Figure{fig:visual-example-1}, \Figure{fig:visual-example-main-2} - \ref{fig:visual-example-main-3} and the supplementary material for visualizations of \airloc{} and \emph{Priv local}.
\\ \\
\textbf{Human performance evaluation.} The results of the proof-of-concept human performance evaluation in \Figure{fig:result_diff_simple} (left) indicate that our proposed task is in general difficult, since only slightly above half of all human controlled trajectories are successful. We also see that \airloc{} and \emph{Priv local} achieve significantly higher success rates compared to human operators. Details about the human performance evaluation are found in the supplementary material.

\subsection{Ablation Study: Motivating the Exploitation Prior}\label{sec:ablations}
In \Figure{fig:result_diff_simple} we evaluate the various \airloc{} variants described earlier,\footnote{Please see the supplement for more ablation results.} together with the best non-RL-based model \emph{Priv local} and the human baseline. \airloc{} is better than its ablated variants on average in both settings ($5 \times 5$ and $7 \times 7$), as well as for most start-to-goal distances (exception at distance 4 in the $7 \times 7$ setting). This motivates the design choice of fully utilizing the exploitation prior within the policy architecture -- see also \Table{table:valdiscrete_noncorrupt_seed}.
\\ \\
Recall that \emph{Priv local} is trained solely in the setting where the start and goal are adjacent, so it can be interpreted as an 'exploitation only' model, where the action distribution is obtained by feeding the exploitation prior $\bs{u}_t$ through a softmax, cf.~\Figure{fig:DetailedPolicy}. Conversely, the \emph{no prior} variant of \airloc{} is trained without any exploitation prior, so the policy must simultaneously learn to explore (search for the goal when it is further away) and exploit (move to the goal when it is adjacent), which may be ambiguous. As seen in \Figure{fig:result_diff_simple}, the \emph{no residual} variant, which allows $\bs{u}_t$ to  guide the agent's decision making by feeding $\bs{u}_t$ to the temporal unit, is only marginally better. Our full \airloc{} agent, which clearly outperforms the other variants, takes this a step further by decoupling exploration and exploitation and only has to learn a residual between the two (since $\bs{u}_t$ is added within the softmax of the decision unit). Hence, during RL training \airloc{} essentially learns when to explore and when to exploit.

\begin{table}[t]
\centering
\caption{Seed sensitivity analysis of the various \airloc{} variants on the validation set of \emph{Massachusetts Buildings} (search area size $5 \times 5$, movement budget $T = 10$). The results on the first lines of each block are the median-performing \airloc{} models and are the ones we have evaluated in the rest of the paper. None of the \airloc{} variants are sensitive to the random seed used for policy network initialization. The worst performing seed of the \emph{no residual} variant of \airloc{} performs better than the best performing seed of the \emph{no prior} variant, and it is also somewhat better than the alternative learnable approach \emph{Priv local}. Similarly, the worst performing seed of our full \airloc{} outperforms the best performing seed of both the ablated variants and \emph{Priv local}, which again motivates our design choices.}\label{table:valdiscrete_noncorrupt_seed}
\vspace{6pt}
\scalebox{0.99}{
\begin{tabular}{|c||c|c|c|c|}
\hline
\textbf{Agent type} & \textbf{Success} & \textbf{Step ratio} & \textbf{Steps} & \textbf{Res. dist.} \\ \hline \hline
\textbf{\airloc{}} & 72.6 \% & 1.49 & 6.0 & 2.4 \\ \hline
\textbf{\airloc{} (other seed \#1)} & 72.2 \% & 1.45 & 6.1 & 2.4 \\ \hline
\textbf{\airloc{} (other seed \#2)} & 72.2 \% & 1.51 & 6.2 & 2.5 \\ \hline
\textbf{\airloc{} (other seed \#3)} & 74.3 \% & 1.56 & 6.2 & 2.4 \\ \hline
\textbf{\airloc{} (other seed \#4)} & 75.9 \% & 1.53 & 6.1 & 2.5 \\ \hline
\textbf{\airloc{} (average)} & 73.4 \% & 1.51 & 6.1 & 2.5 \\ \hline
\hline
\textbf{\airloc{} (no residual)} & 68.5 \% & 1.49 & 6.3 & 2.2 \\ \hline
\textbf{\airloc{} (no residual, other seed \#1)} & 68.6 \% & 1.52 & 6.3 & 2.2 \\ \hline
\textbf{\airloc{} (no residual, other seed \#2)} & 69.5 \% & 1.52 & 6.3 & 2.2 \\ \hline
\textbf{\airloc{} (no residual, other seed \#3)} & 68.2 \% & 1.60 & 6.4 & 2.2 \\ \hline
\textbf{\airloc{} (no residual, other seed \#4)} & 67.2 \% & 1.57 & 6.4 & 2.2 \\ \hline
\textbf{\airloc{} (no residual, average)} & 68.4 \% & 1.54 & 6.3 & 2.2 \\ \hline
\hline
\textbf{AiRLoc (no prior)}      &  65.9 \% & 1.56 & 6.5  & 2.4 \\ \hline
\textbf{AiRLoc (no prior, other seed \#1)}  &  64.8 \% & 1.56 & 6.7  & 2.4 \\ \hline
\textbf{AiRLoc (no prior, other seed \#2)}  &  66.6 \% & 1.56 & 6.5  & 2.5 \\ \hline
\textbf{AiRLoc (no prior, other seed \#3)}  &  66.6 \% & 1.50 & 6.4  & 2.3 \\ \hline
\textbf{AiRLoc (no prior, other seed \#4)}  &  64.9 \% & 1.50 & 6.6  & 2.4 \\ \hline
\textbf{\airloc{} (no prior, average)} & 65.8 \% & 1.54 & 6.5 & 2.4 \\ \hline
\hline
\textbf{Priv local}  & 67.0 \% & 1.54 & 6.3 & 2.3 \\ \hline
\end{tabular}}
\vspace{-8pt}
\end{table}

\subsection{Random Seed Sensitivity Analysis}\label{sec:seed-analysis}
\Table{table:valdiscrete_noncorrupt_seed} shows the results of a seed sensitivity analysis (regarding policy network initilization) for \airloc{} and its ablated variants on the validation set of \emph{Massachusetts Buildings}. The \airloc{} variants are trained with five random network initializations each until convergence on the validation set, and the results for the median-performing models on the validation set are the ones reported within the rest of the paper. The seed sensitivity is low overall. Furthermore, our full \airloc{} agent outperforms \emph{Priv local} even for the worst-performing seed.

\section{Conclusions}
In this work we have introduced the novel \emph{aerial view goal localization} task and framework, which allows for controllable and reproducible development of methodologies that can eventually be useful for automated search-and-rescue operations, e.g.~in regions that are heavily affected by climate-induced disasters. Naturally, as with most technologies, there are also possible applications that may be unethical. We strongly discourage extending our research in such directions, and instead call for extensions towards benign use-cases.
\\ \\
The difficulty for humans to perform well on our proposed task shows that it is a reasonable first step for model development and evaluation, even though the setup avoids some challenges of real use-cases. Relevant next steps toward making the proposed methodologies more practically useful include making the goal specification more flexible (e.g.~allowing for a ground-level image description of the goal); requiring the agent to explicitly declare when it has reached its goal; and considering even larger search areas. 
\\ \\
A reinforcement learning-based approach, \emph{\airloc{}}, was developed to tackle the proposed task, in addition to several other learnable and heuristic methods. Key components of the policy architecture include a mid-level vision module and an explicit decoupling between exploration and exploitation, both of which were shown to be crucial for \airloc{}'s performance. Extensive experimental evaluations clearly showed the benefits of our \airloc{} agent over the learnable and heuristic baselines. In particular, our methodology can be used to localize goals in aerial images depicting disaster zones, despite being trained only on scenarios without disasters. Code and models have been made publicly available\footnote{\url{https://github.com/aleksispi/airloc}} so that others can further explore and extend our proposed task towards real use-cases, for example within disaster relief and management. 
\clearpage


\newpage
\bibliography{clean-arxiv}
\bibliographystyle{iclr2023_conference}

\clearpage
\appendix
\section{Supplementary Material}



\noindent In this supplementary material we provide additional results and insights for our proposed \airloc{} agent, baselines and the datasets we have used. In \Section{sec:sm-viz} we provide several additional qualitative examples (visualizations) of \airloc{} and the second best approach \emph{Priv local}. In \Section{sec:sm-details} we provide more details about the policy architecture of \airloc{}. An extended ablation study is presented in \Section{sm:ablation}. Further dataset details are given in \Section{sec:sm-dataset}. Finally, a description of the human performance evaluation is found in \Section{sm:human}.

\subsection{Visualizations of Agent Trajectories}\label{sec:sm-viz}
\begin{figure}
    \centering
         \includegraphics[width=.985
         \textwidth]{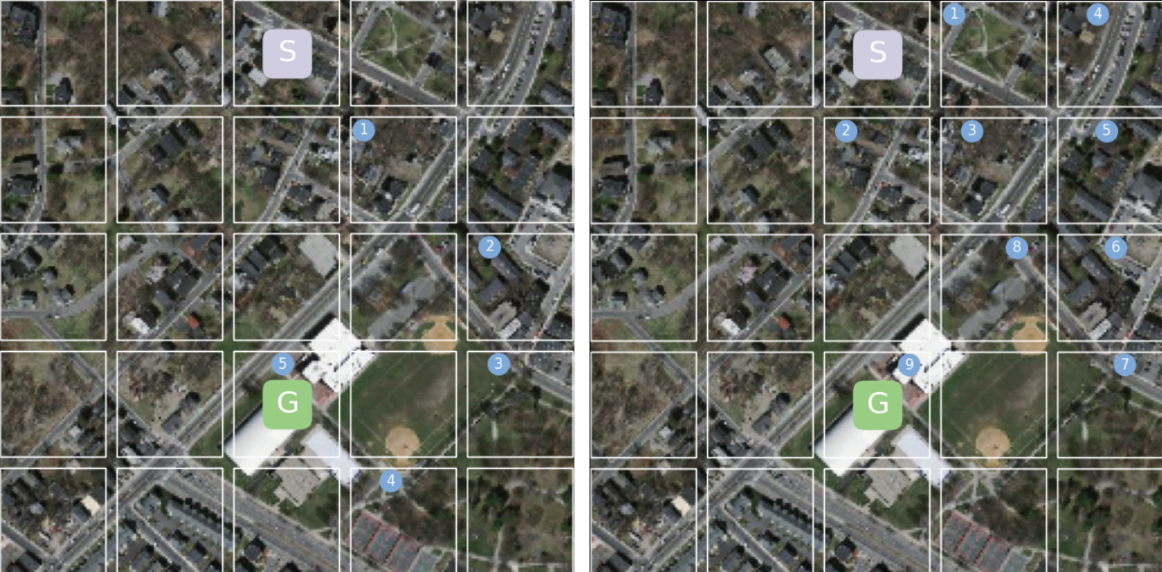}
    \vspace{-6pt}
    \caption{Successful examples of \airloc{} (left) and \emph{Priv local} (right) on the \emph{Masa} test set ($5 \times 5$ setup, movement budget $T=10$). The start and goal locations are denoted 'S' and 'G', respectively. The numbered circles show which locations are visited and in what order. Recall that the full underlying search area is never observed in its entirety (they are shown here for visualization purposes only), i.e.~the agents must operate based on partial glimpses alone. \airloc{} takes a different and much shorter path towards the goal location.}
    \label{fig:visual-example-11-masa}
\end{figure}

\begin{figure}[h!]
    \centering
         \includegraphics[width=.985
         \textwidth]{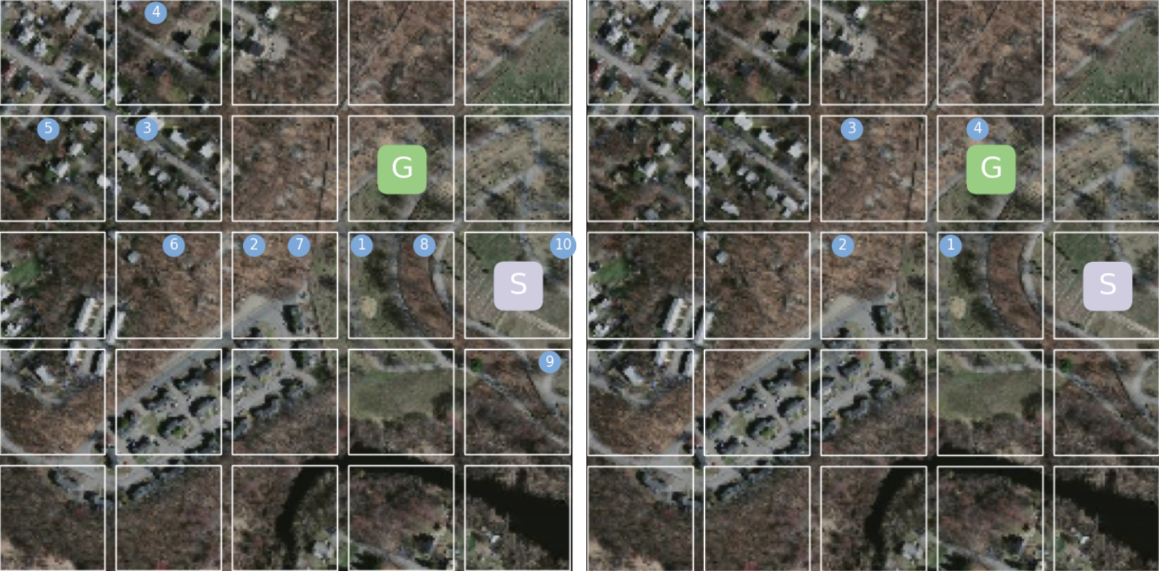}
    \vspace{-4pt}
    \caption{An unsuccessful example of \airloc{} (left) and a successful example of \emph{Priv local} (right) on the \emph{Masa} test set ($5 \times 5$ setup, movement budget $T=10$). \airloc{} moves in the wrong direction early on, even though it manages to backtrack and get close to the goal again (e.g.~location \#1 and \#8 coincide). However, \airloc{} ultimately fails to find the goal location in this example.}
    \label{fig:visual-example-14-masa}
\end{figure}

\begin{figure}[h!]
    \centering
         \includegraphics[width=.985
         \textwidth]{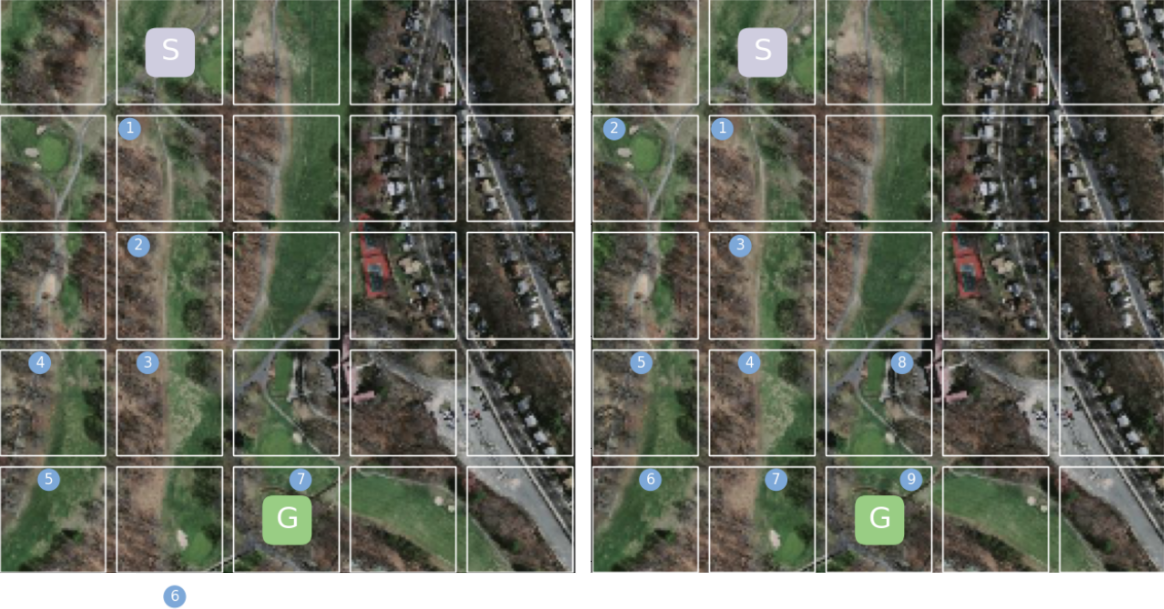}
    \vspace{-4pt}
    \caption{Successful examples of \airloc{} (left) and \emph{Priv local} (right) on the \emph{Masa} test set ($5 \times 5$ setup, movement budget $T=10$). \airloc{} and \emph{Priv local} take the same first action, then \airloc{} deviates from the exploitation prior and takes a shorter path towards the goal. Note that \airloc{} even moves outside the search area but still reaches the goal. Recall that \emph{Priv local} has explicit restrictions which ensure that it always stays within the search area (as shown in the main paper, without such privileges this approach yields very poor results).}
    \label{fig:visual-example-2-masa}
\end{figure}

\begin{figure}
    \centering
         \includegraphics[width=.985
         \textwidth]{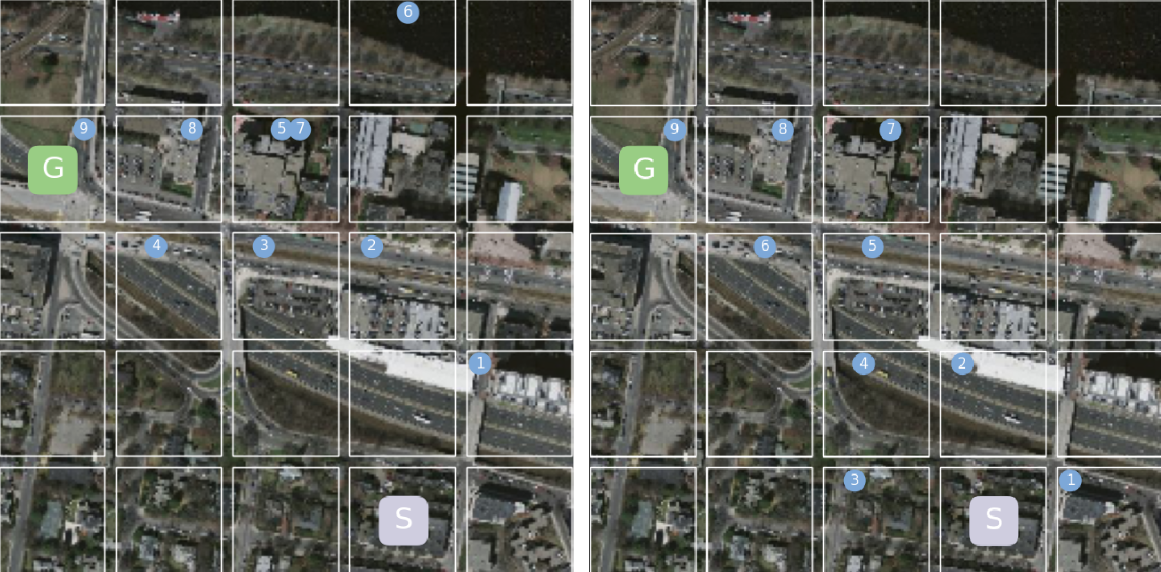}
    \vspace{-4pt}
    \caption{Successful examples of \airloc{} (left) and \emph{Priv local} (right) on the \emph{Masa} test set ($5 \times 5$ setup, movement budget $T=10$). In this example \airloc{} begins by deviating from the exploitation prior and explores the area differently. Note in particular how it takes a suboptimal action from location \#5 to location \#6 (instead of moving left towards the goal), then recovers and backtracks (location \#5 and \#7 coincide), and finally resorts to the exploitation prior (compare \#7 - \#9 with \emph{Priv local} on the right) which takes it to the goal location. Both agents reach the goal location in the same number of steps.}
    \label{fig:visual-example-9-masa}
\end{figure}

\begin{figure}
    \centering
         \includegraphics[width=.985
         \textwidth]{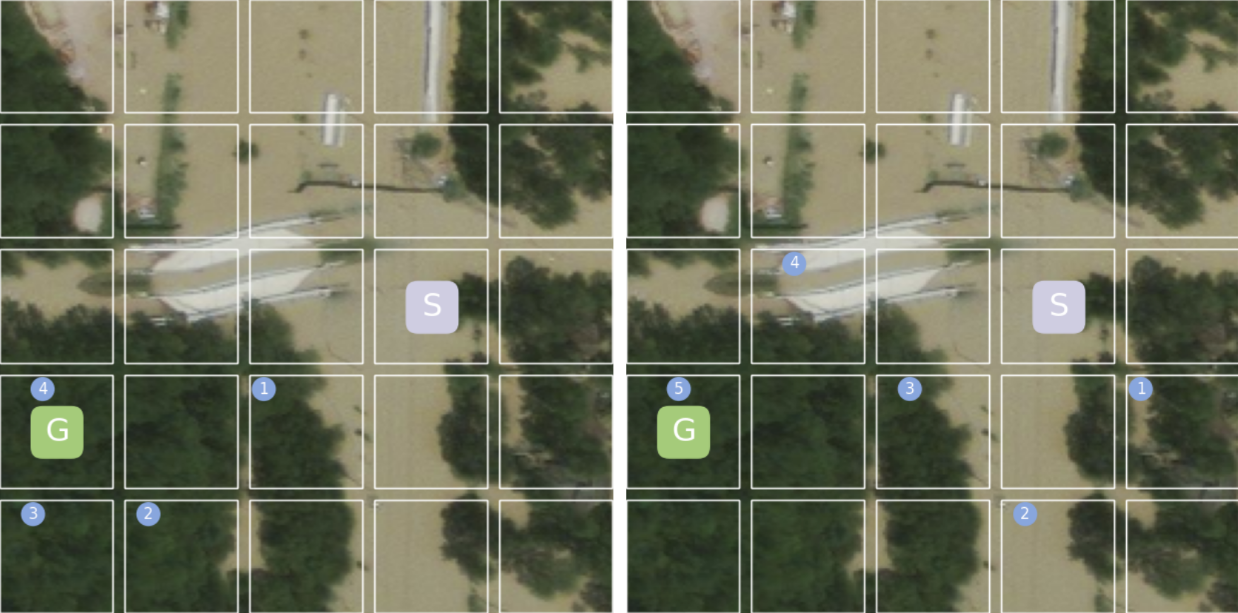}
    \vspace{-4pt}
    \caption{Successful examples of \airloc{} (left) and \emph{Priv local} (right) on a flooding scenario in \emph{xBD-disaster} ($5 \times 5$ setup, movement budget $T=10$). \airloc{} takes a different and slightly shorter path towards the goal location.}
    \label{fig:visual-example-14}
\end{figure}

\begin{figure}[h!]
    \centering
         \includegraphics[width=.985
         \textwidth]{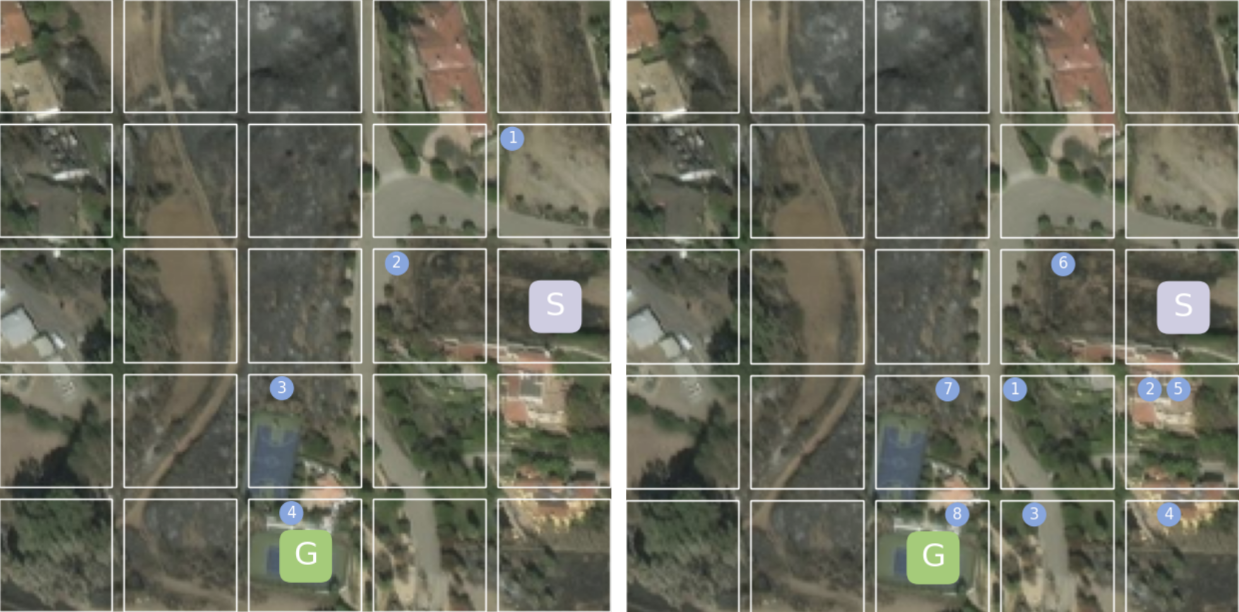}
    \vspace{-4pt}
    \caption{Successful examples of \airloc{} (left) and \emph{Priv local} (right) on a post-wildfire scenario in \emph{xBD-disaster} ($5 \times 5$ setup, movement budget $T=10$). \airloc{} takes a different and significantly shorter path towards the goal location. Note that \emph{Priv local} visits the location below the start location after 2 and 5 steps, despite its privileged movement constraints which tries to avoid previous locations. However, in this example, after the 4th step there are no unvisited locations to move to, and so it has to move somewhere.}
    \label{fig:visual-example-2}
\end{figure}

\begin{figure}
    \centering
         \includegraphics[width=.985
         \textwidth]{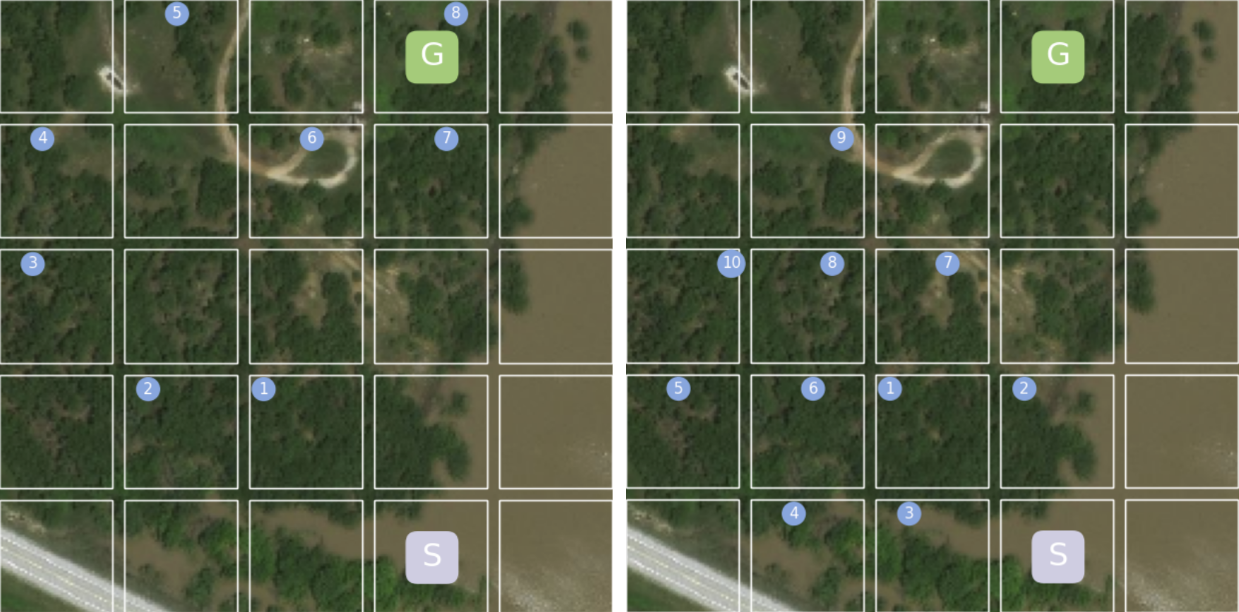}
    \vspace{-4pt}
    \caption{A Successful example of \airloc{} (left) and an unsuccessful example of \emph{Priv local} (right) on a flooding scenario in \emph{xBD-disaster} ($5 \times 5$ setup, movement budget $T=10$). \airloc{}'s location \#7 shares forest-structure with the goal location, which may have been an important visual cue in the last step.}
    \label{fig:visual-example-3}
\end{figure}

\begin{figure}[h!]
    \centering
         \includegraphics[width=.985
         \textwidth]{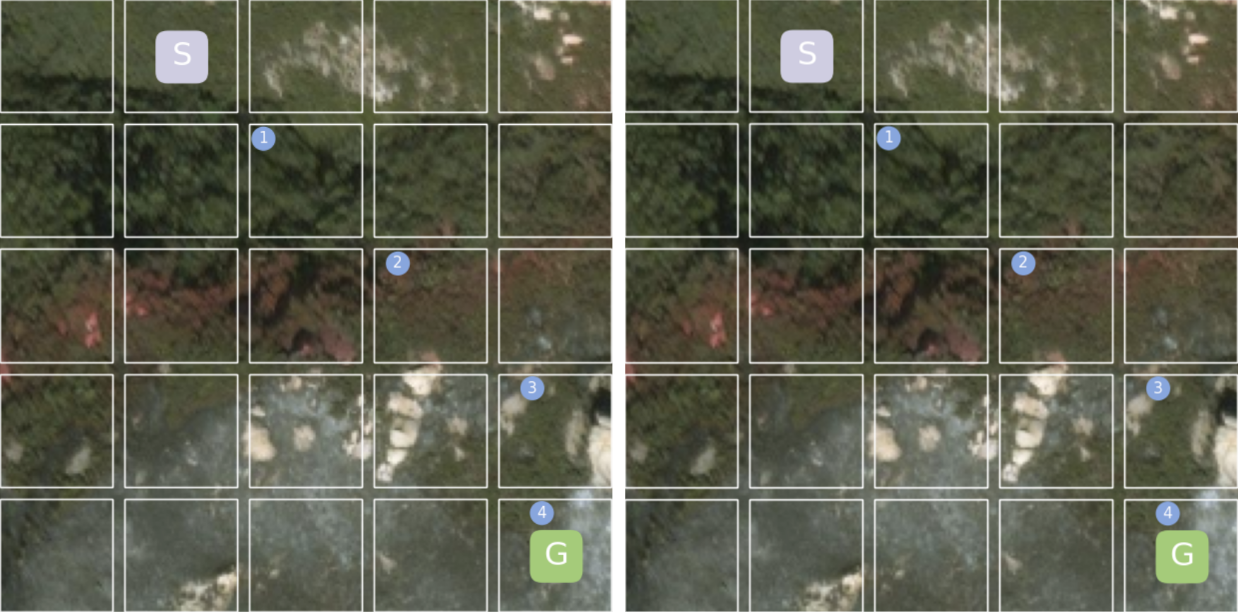}
    \vspace{-4pt}
    \caption{Successful examples of \airloc{} (left) and \emph{Priv local} (right) on a wildfire scenario in \emph{xBD-disaster} ($5 \times 5$ setup, movement budget $T=10$). Both agents take the exact same (and shortest) path towards the goal, i.e.~\airloc{} fully resorts to the exploitation prior in this case.}
    \label{fig:visual-example-4}
\end{figure}

\begin{figure}
    \centering
         \includegraphics[width=.985
         \textwidth]{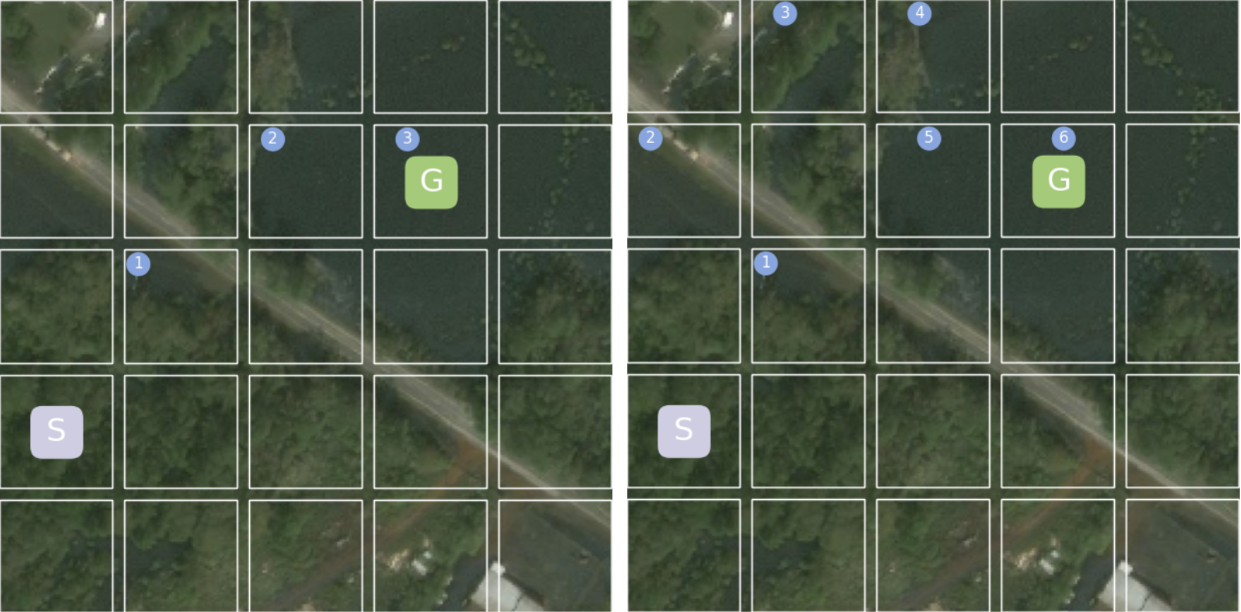}
    \vspace{-4pt}
    \caption{Successful examples of \airloc{} (left) and \emph{Priv local} (right) on a flooding scenario in \emph{xBD-disaster} ($5 \times 5$ setup, movement budget $T=10$). \airloc{} takes the first same step as \emph{Priv local}, then deviates and takes a shortest path towards the goal. \emph{Priv local} reaches the goal using several more steps.}
    \label{fig:visual-example-5}
\end{figure}

\begin{figure}[h!]
    \centering
         \includegraphics[width=.985
         \textwidth]{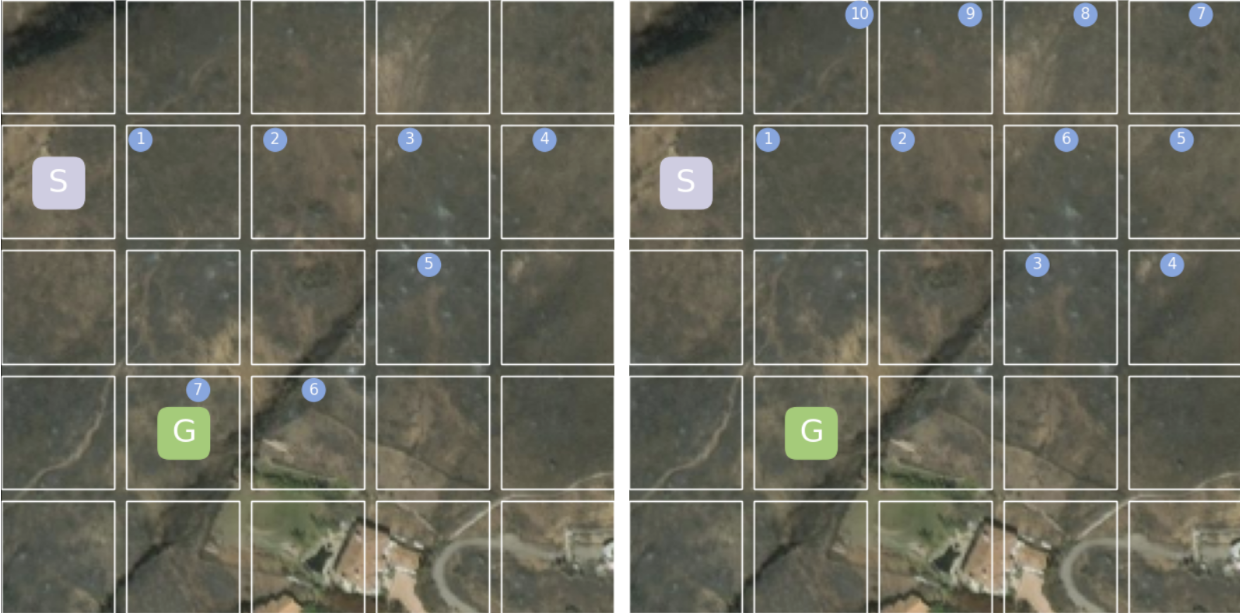}
    \vspace{-4pt}
    \caption{A successful example of \airloc{} (left) and an unsuccessful example of \emph{Priv local} (right) on a post-wildfire scenario in \emph{xBD-disaster} ($5 \times 5$ setup, movement budget $T=10$). \airloc{} moves in the same way as \emph{Priv local} for the first two steps and then deviates. Note that \airloc{} does not take the shortest path towards the goal but nonetheless reaches it well within the movement budget.}
    \label{fig:visual-example-6}
\end{figure}

\begin{figure}
    \centering
         \includegraphics[width=.985
         \textwidth]{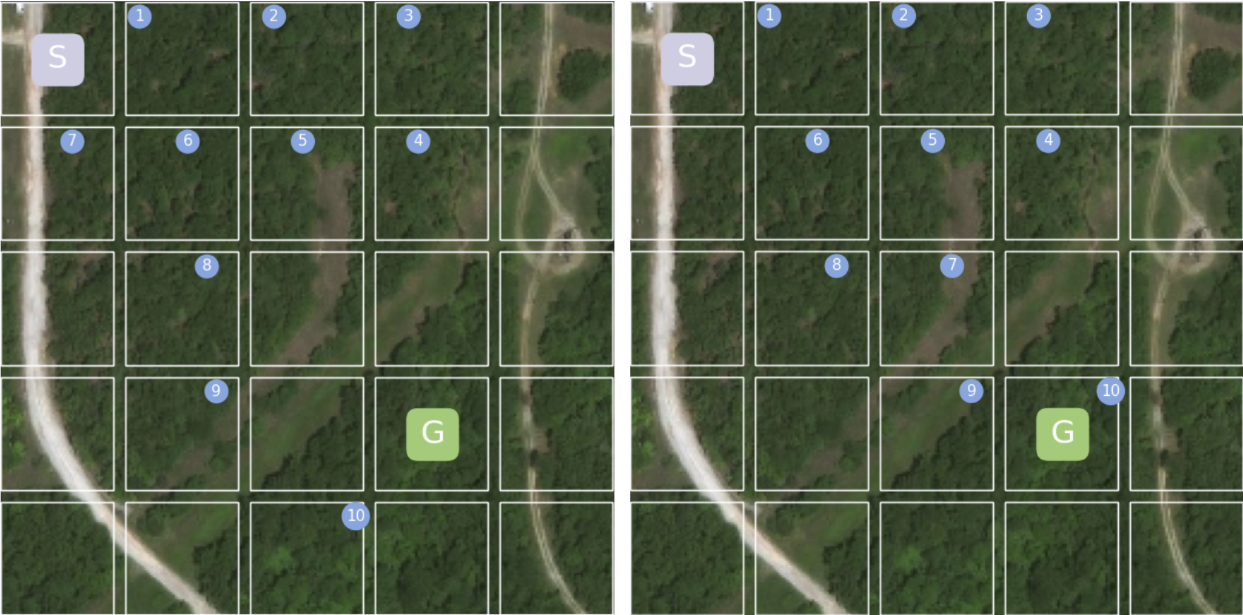}
    \vspace{-4pt}
    \caption{An unsuccessful example of \airloc{} (left) and a successful example of \emph{Priv local} (right) on \emph{xBD-disaster} ($5 \times 5$ setup, movement budget $T=10$). \airloc{} takes the same path as \emph{Priv local} for the first six steps and then deviates (it is adjacent to the goal when the budget $T=10$ is exhausted). \emph{Priv local} precisely manages to reach the goal within the budget.}
    \label{fig:visual-example-15}
\end{figure}

\begin{figure}[h!]
    \centering
         \includegraphics[width=.985
         \textwidth]{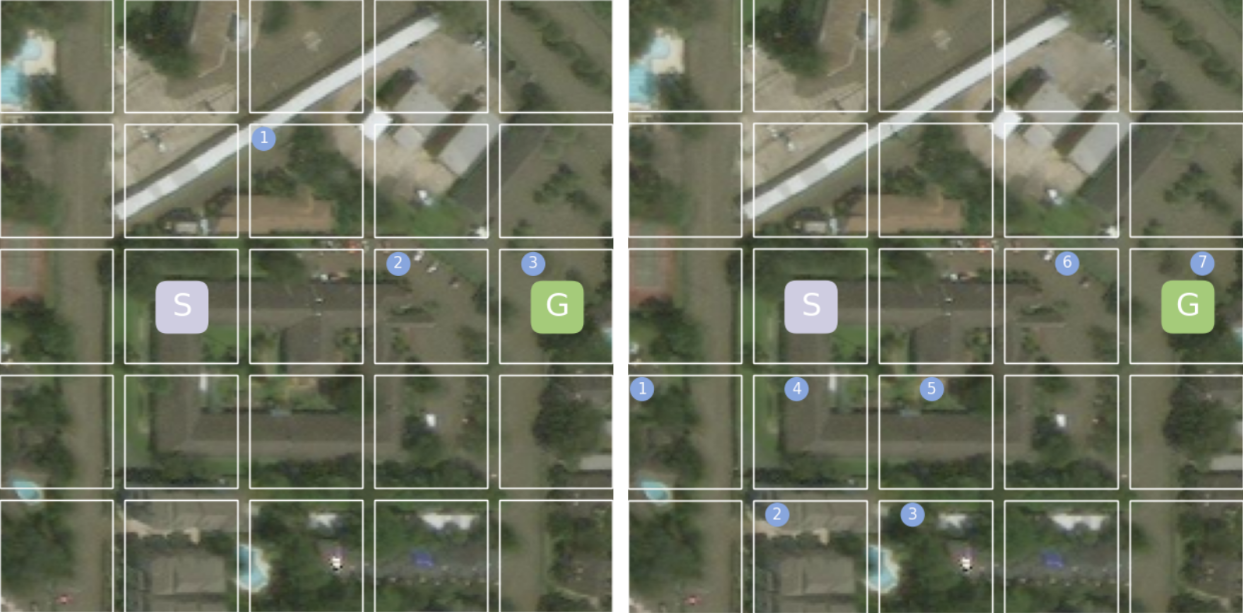}
    \vspace{-4pt}
    \caption{Successful examples of \airloc{} (left) and \emph{Priv local} (right) on a flooding scenario in \emph{xBD-disaster} ($5 \times 5$ setup, movement budget $T=10$). \airloc{} takes a different and much shorter path towards the goal location.}
    \label{fig:visual-example-16}
\end{figure}

\begin{figure}
    \centering
         \includegraphics[width=.985
         \textwidth]{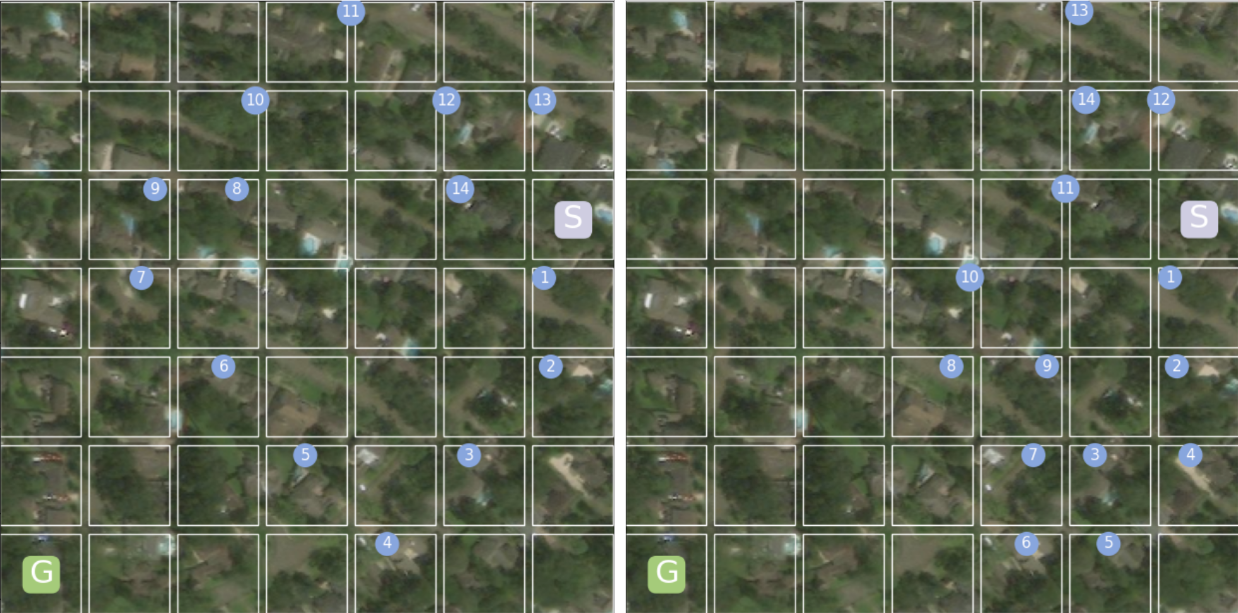}
    \vspace{-4pt}
    \caption{Unsuccessful examples of \airloc{} (left) and \emph{Priv local} (right) on a flooding scenario in \emph{xBD-disaster} ($7 \times 7$ setup, movement budget $T=14$). \airloc{} takes the same first two steps as \emph{Priv local}, then deviates, but (like \emph{Priv local}) fails to reach the goal.}
    \label{fig:visual-example-8}
\end{figure}

\begin{figure}[h!]
    \centering
         \includegraphics[width=.985
         \textwidth]{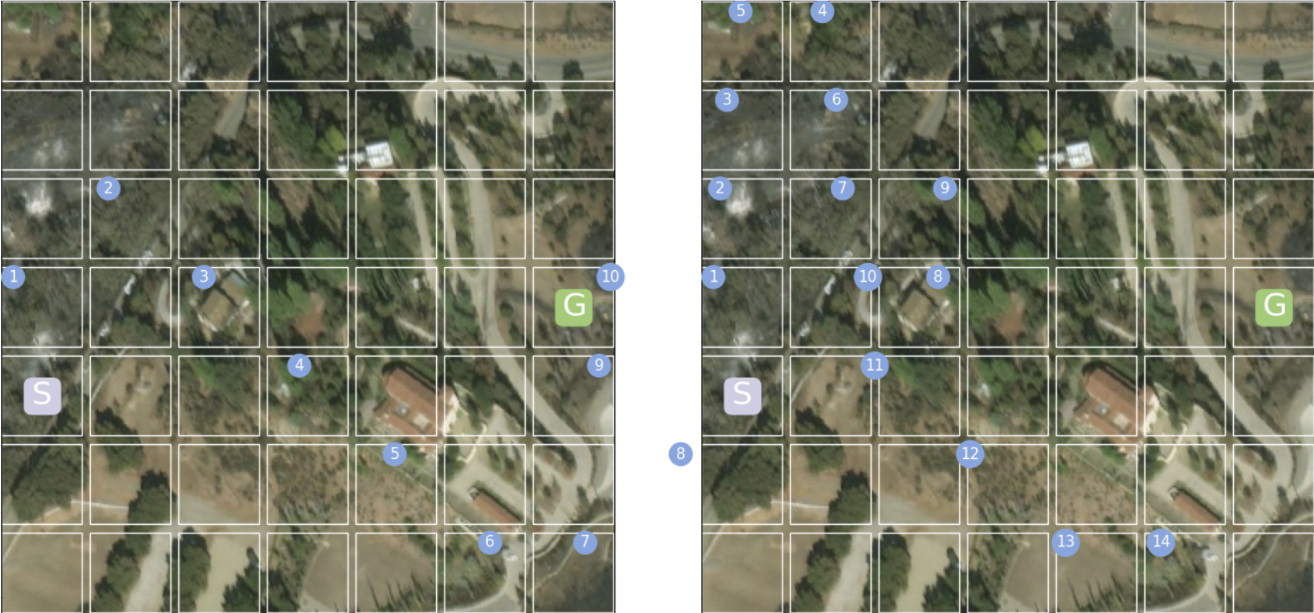}
    \vspace{-4pt}
    \caption{A successful example of \airloc{} (left) and an unsuccessful example of \emph{Priv local} (right) on \emph{xBD-disaster} ($7 \times 7$ setup, movement budget $T=14$). \airloc{} takes the same first step as \emph{Priv local} and then deviates. Note that \airloc{} even moves outside the search are at one occasion (location \#8), but still manages to reach the goal well within the movement budget.}
    \label{fig:visual-example-9}
\end{figure}

In Fig. \ref{fig:visual-example-11-masa} - \ref{fig:visual-example-9-masa} we show additional qualitative examples of AiRLoc and the best alternative learnable approach \emph{Priv local} on the test set of \emph{Massachusetts Buildings} (\emph{Masa}). In this case the models were trained on the training set of \emph{Masa}. Fig. \ref{fig:visual-example-14} - \ref{fig:visual-example-9} show additional visualizations of \airloc{} and \emph{Priv local} on disaster-hit search areas from the dataset \emph{xBD-disaster}. In this case the models were trained on non-disaster-hit data from \emph{xBD-pre}, where we have ensured that this training data depicts other geographical areas than those in \emph{xBD-disaster}. See more detailed information about each dataset in \Section{sec:sm-dataset}.
\\ \\
When inspecting these visual examples, keep in mind the connection between \airloc{} and \emph{Priv local}, where \emph{Priv local} is essentially an 'exploit only' model that is optimized to localize adjacent goals. The action distribution of \emph{Priv local} is obtained\footnote{Subject to privileged movement constraints, without which it performs abysmally (see \Table{table:valdiscrete_noncorrupt_simple}).} by feeding its final output $\bs{u}_t \in \mathbb{R}^8$ through a softmax. Our full \airloc{} agent takes advantage of this exploitation prior $\bs{u}_t$ and decouples exploration from exploitation, as explained in the main paper. \airloc{} thus decides when to resort to \emph{Priv local}'s exploitative behavior (although without the privileges) and when to explore independently.

\clearpage




\clearpage


\clearpage


\clearpage


\clearpage


\clearpage

\subsection{More Details About the Patch Embedder and Segmentation Unit}\label{sec:sm-details}
Pretraining backbone vision components is common in RL setups, since it often yields a higher end performance \citep{sax2018mid,visionPretrain,wang2022vrl3,xiao2022masked,yadav2022offline}. Before training the rest of \airloc{} with reinforcement learning, the patch embedder is therefore pretrained (in the same self-supervised fashion as \cite{doerch}) on the training set of \emph{Massachusetts Buildings} (or on \emph{xBD-pre}, depending on which experiment is considered) using the categorical cross-entropy loss. This loss is computed using the 8-dimensional patch embedder prediction and a one hot encoding of the true goal direction relative to the start location (recall that during this pretraining stage, the start and goal are assumed to be adjacent). The Adam optimizer \citep{kingma2014adam} with batches of 256 pairs of image patches (start and goal) and a learning rate of $10^{-3}$ is used during this pretraining phase.
\\ \\
For the segmentation unit, we use and adapt the U-net model for biomedical segmentation applications \citep{UNet}. A publicly available implementation of this U-net\footnote{\url{https://github.com/milesial/Pytorch-UNet}} is used as a starting point. However, since the patches (partial glimpses of the search area) are smaller than in the original U-net, the network structure is altered. This altered network consist of four downsampling convolutional blocks, which reduce the spatial dimensions of the input into a $3 \times 3 \times 64$ embedding. Then, four upsampling convolutional blocks are used to recreate the spatial dimension of the input patch (thus the segmentation unit outputs a binary $48 \times 48 \times 1$ building segmentation mask, although in general the segmentation unit could obviously include more classes as well). The segmentation network is pretrained on the training set of \emph{Massachusetts Buildings} using a cross-entropy loss with Adam, batch size 128, and learning rate $10^{-4}$, and is kept frozen when training the rest of \airloc{}.

\subsection{Extended Ablation Study}\label{sm:ablation}
See \Table{table:valdiscrete_noncorrupt_simple} for an extended ablation study of \airloc{} on the validation set of \emph{Massachusetts Buildings}. For convenience, we here repeat the definitions of the various \airloc{} variants. \textbf{\emph{No sem seg}} omits the segmentation unit and uses only RGB patches in the patch embedder (which is instead pretrained with RGB-only inputs). \textbf{\emph{No residual}} omits $\bs{u}_t$ in the decision unit, but not in the temporal unit, cf.~\Figure{fig:DetailedPolicy} in the main paper. \textbf{\emph{No prior}} entirely discards the prior $\bs{u}_t$ in the architecture. Finally, \textbf{\emph{Priv}} refers to the use of the privileged movement constraints which i) ensures that the agent cannot move outside the search area; and ii) it avoids previous locations.
\begin{table}[ht!]
\centering
\caption{Extended ablation study on the validation set of \emph{Massachusetts Buildings} (movement budget $T=10$ and $T=14$ for setups of sizes $5 \times 5$ and $7 \times 7$, respectively). Adding the movement constraint privileges of \emph{Priv local} does not yield any significant improvements for \airloc{} -- it even reduces the success rate for our full \airloc{} agent. Conversely, in the bottom of this table we report results for \emph{Local}, which is the same as \emph{Priv local} but without the privileged movement constraints (thus \emph{Local} may visit the same location multiple times and move outside the search area). Different to \airloc{}, which also lacks any privileged movement constraints, \emph{Local} performs abysmally when it is not given such constraints. Mid-level vision capabilities (semantic segmentation) are crucial for \airloc{}'s performance. The fact that the ablated \airloc{} variants generally result in a lower success rate motivates our design choices.}\label{table:valdiscrete_noncorrupt_simple}
\vspace{6pt}
\scalebox{0.95}{
\begin{tabular}{|c||c|c|c|c|}
\hline
\textbf{Agent type} & \textbf{Success} & \textbf{Step ratio} & \textbf{Steps} & \textbf{Res. dist.} \\ \hline \hline
\textbf{\airloc{} (5x5)} & 72.6 \% & 1.49 & 6.0 & 2.4 \\ \hline
\textbf{\airloc{} (priv, 5x5)} & 68.8 \% & 1.47 & 6.2 & 2.3 \\ \hline
\textbf{\airloc{} (no residual, 5x5)} & 68.5 \% & 1.49 & 6.3 & 2.2 \\ \hline
\textbf{\airloc{} (no residual, priv, 5x5)} & 71.9 \% & 1.52 & 6.2 & 2.5 \\ \hline
\textbf{AiRLoc (no prior, 5x5)}            &  65.9 \% & 1.56 & 6.5  & 2.4 \\ \hline
\textbf{AiRLoc (no prior, priv, 5x5)}            &  67.1 \% & 1.56 & 6.4  & 2.6 \\ \hline
\hline
\textbf{\airloc{} (no sem seg, 5x5)} & 62.6 \% & 1.52 & 6.6 & 2.4 \\ \hline
\textbf{\airloc{} (no sem seg, priv, 5x5)} & 64.6 \% & 1.56 & 6.7 & 2.5 \\ \hline
\textbf{\airloc{} (no residual, no sem seg, 5x5)} & 61.1 \% & 1.56 & 6.8 & 2.4 \\ \hline
\textbf{\airloc{} (no residual, no sem seg, priv, 5x5)} & 62.6 \% & 1.59 & 6.8 & 2.5 \\ \hline
\textbf{AiRLoc (no prior, no sem seg, 5x5)} & 60.7 \% & 1.67 & 6.9  & 2.5 \\ \hline
\textbf{AiRLoc (no prior, no sem seg, priv, 5x5)} &  62.2 \%       & 1.69 & 6.9  & 2.6 \\ \hline
\hline
\textbf{\airloc{} (7x7)} & 57.6 \% & 1.54 & 9.6 & 3.4 \\ \hline
\textbf{\airloc{} (priv, 7x7)} & 51.4 \% & 1.49 & 10.1 & 3.3 \\ \hline
\textbf{\airloc{} (no residual, 7x7)} & 50.5 \% & 1.54 & 10.1 & 3.4 \\ \hline
\textbf{\airloc{} (no residual, priv, 7x7)} & 52.3 \% & 1.54 & 9.9 & 3.5 \\ \hline
\textbf{AiRLoc (no prior, 7x7)}            & 50.5 \% & 1.59 & 10.2 & 3.6 \\ \hline
\textbf{AiRLoc (no prior, priv, 7x7)}      & 52.1 \% & 1.59 & 10.1 & 3.6 \\ \hline
\hline
\textbf{\airloc{} (no sem seg, 7x7)} & 48.4 \% & 1.56 & 10.4 & 3.3 \\ \hline
\textbf{\airloc{} (no sem seg, priv, 7x7)} & 47.7 \% & 1.69 & 10.7 & 3.2 \\ \hline
\textbf{\airloc{} (no residual, no sem seg, 7x7)} & 46.1 \% & 1.59 & 10.4 & 3.6 \\ \hline
\textbf{\airloc{} (no residual, no sem seg, priv, 7x7)} & 48.8 \% & 1.64 & 10.4 & 3.7 \\ \hline
\textbf{AiRLoc (no prior, no sem seg, 7x7)} & 42.4 \% & 1.79 & 11.1 & 3.4 \\ \hline
\textbf{AiRLoc (no prior, no sem seg, priv, 7x7)} & 44.4 \%       & 1.82 & 11.0 & 3.4 \\ \hline
\hline
\textbf{Priv local}  & 67.0 \% & 1.54 & 6.3 & 2.3 \\ \hline
\textbf{Local} &  19.9 \% & 0.79 & 8.5 & 6.1 \\ \hline
\end{tabular}}
\end{table}

\subsection{Dataset Details}\label{sec:sm-dataset}
As described in the main paper, \emph{Massachusetts Buildings} is used as the main dataset for model development and evaluation. There are 832 different search areas in training (70\%), 178 in validation (15\%), and 178 in testing (15\%). Since top-right and left-right flipping of search areas is performed during training, and since search a area of size $M \times N=5 \times 5$ has $25 \cdot 24$ different start-goal configurations, there are in total $832 \cdot 4 \cdot 25 \cdot 24 \approx 2 \cdot 10^6$ unique training configurations. As the various agents are trained for roughly 50k batches each, and since each batch consists of 64 episodes, this amounts to $3.2 \cdot 10^6$ training episodes.
\\ \\
During evaluation, one randomly generated but fixed configuration of each start-to-goal distance is used per search area, which results in 712 fixed validation and test configurations, respectively, in the $5 \times 5$ setting ($4 \cdot 178 = 172$). Similarly, when evaluating on the \emph{Dubai} dataset \citep{dubai}, there are 196 search areas and thus 784 fixed evaluation configurations. The grid cells of the search areas are of size $48 \times 48 \times 3$, with 4 pixels between each other to avoid overfitting models to edge artefacts (each cell corresponds to roughly $100 \times 100$ meters).
\\ \\
As for the \emph{xBD-pre} and \emph{xBD-disaster} data, they again depict data from non-disaster-hit (\emph{xBD-pre}) and disaster-hit (\emph{xBD-disaster}) areas,\footnote{More specifically, \emph{xBD-pre} contains the satellite image subset depicting various locations prior to hurricane Michael (found in the \texttt{tier1} subset of the \emph{xBD} dataset), and \emph{xBD-disaster} contains the satellite image subset depicting various locations after various natural disasters (also found in the \texttt{tier1} subset of the \emph{xBD} dataset).} and the respective data splits are from different geographical areas (thus there is no spatial overlap). There are 902 different search areas in training (\emph{xBD-pre}), and since top-right and left-right flipping of search areas is performed during training, and since search a area of size $M \times N=5 \times 5$ has $25 \cdot 24$ different start-goal configurations, there are in total $902 \cdot 4 \cdot 25 \cdot 24 \approx 2.2 \cdot 10^6$ unique training configurations. During evaluation (on \emph{xBD-disaster}), one randomly generated but fixed configuration of each start-to-goal distance is used per search area, which results in 5212 evaluation configurations in the $5 \times 5$ setting (there are 1303 search areas in \emph{xBD-disaster} and $4 \cdot 1303 = 5212$).

\section{Description of the Human Performance Evaluation}\label{sm:human}
\begin{figure}[t]
    \centering
    \includegraphics[width = 0.97 \textwidth]{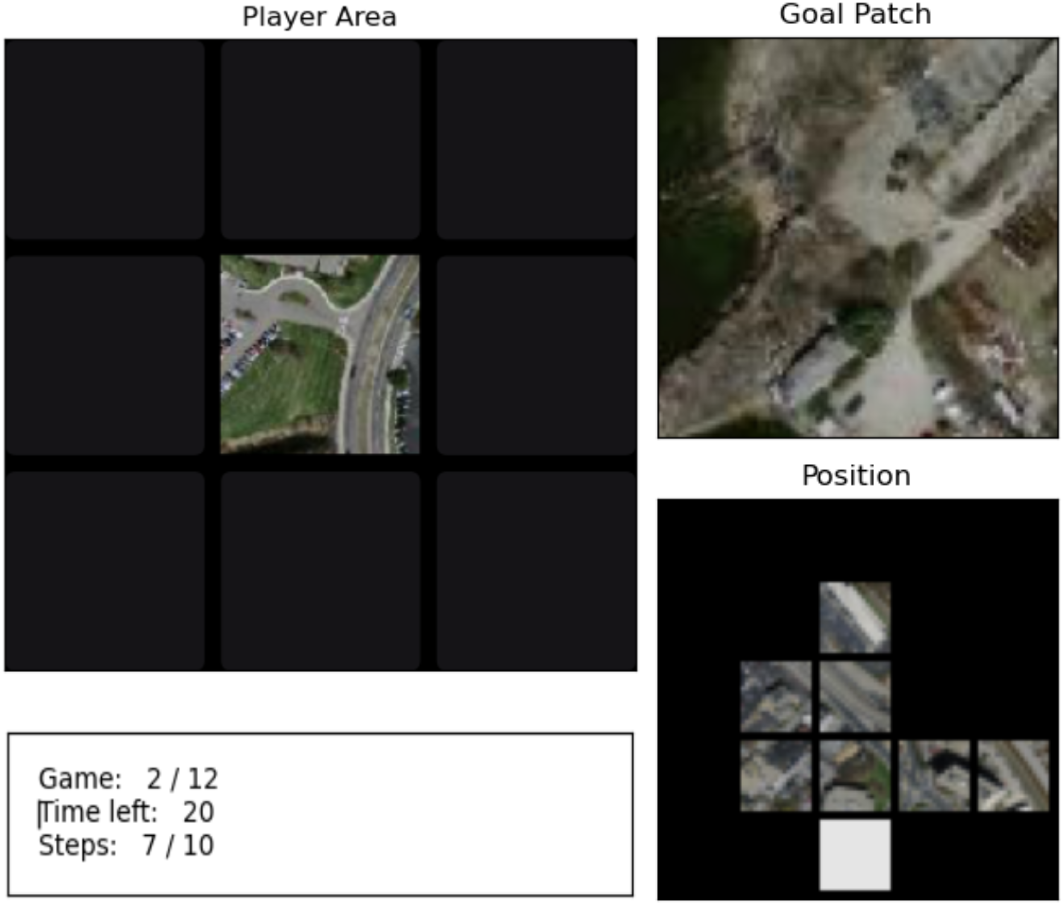}
    \vspace{-6pt}
    \caption{An example of the human performance evaluation setup. Each participant was given a set of 12 different such games (a game is a search area and an associated start and goal location), and there was no overlap in the games played by different participants. Each search area was of size $5 \times 5$ and the movement budget was $T=10$.}\label{fig:HumanGame}
    \vspace{-6pt}
\end{figure}

To compare the performance of \airloc{} with a human operator in a similar setting, a game version of the task was developed. For fair comparisons, this game was designed to resemble how \airloc{} perceives the search area. Therefore, in addition to receiving the current and goal patches, the human operator is also aware of the borders of the search area, and knows the current position as well as the history of all previously visited positions within the confined area -- see \Figure{fig:HumanGame}. In fact, the human operator can even see all the previously visited patches, while this information is not provided to AiRLoc. We decided to provide humans with this additional information as they have not been trained for the task at hand.  Based on this input, the human operator can move to any of the eight adjacent patches. The movement is selected by clicking with a mouse cursor on one of the eight dark squares surrounding the current location in the \emph{Player Area}, shown on the left in \Figure{fig:HumanGame}. The game uses search areas of size $5 \times 5$ and ends either when the movement budget $T=10$ is exhausted or when the player moves into the goal location (just as for \airloc{} and the other baselines). Moreover, different to the other approaches, the human participants have a limited time to complete each game (60 seconds). Such a time limit was used for the convenience of the participants -- we wanted to avoid that the participants felt like they had to spend several minutes per action to squeeze out the maximum possible performance. The 60 second time limit was assessed to be more than sufficient for completing each game, and the participants agreed with this. 
\vspace{9.2pt}

\noindent The age span of the 19 people who participated is between 14 and 42 years, with an average of 26.4 years and a median of 25 years. There were 13 men and 6 women (68\% and 32\%, respectively). For each human operator, 12 unique search areas from the validation set of \emph{Massachusetts Buildings} were used, as well as a few sample search areas for the player to get acquainted with the controls of the game -- the participants were able to practice as long as they desired, and no statistics were tracked during this warm up phase. The exact games provided span a subset of the games that \airloc{} and the other baselines are evaluated on, to ensure that the comparison is as fair as possible. However, each human is not tested on the entire dataset since it is impractically large, and hence there is a higher uncertainty in the human performance evaluation. The difficulty settings were split equally over these twelve games, with three games per difficulty (here difficulty is the distance between the start and goal patches, ranging from 1 to 4 steps away).
\\ \\
Even though the human setup is very similar to that of \airloc{}, there are some concepts that do not translate well to a human controlled setup. First, the positional encoding of \airloc{} is difficult to translate to human visual processing, and instead a map of the positions was implemented (thus the participants receive explicit information from past locations, different from \airloc{}). Second, the human participants have not trained on the task like \airloc{}, and their visual systems are likely not tailored towards handling the quite low resolution patches. On the other hand, humans have implicitly conducted a lifetime worth of generic visual pretraining, which AiRLoc has not. These discrepancies, in conjunction with the limited number of human controlled trajectories, somewhat limit the reliability of the human baseline. Nonetheless, it is still a useful indication of the human performance on our proposed task.

\end{document}

%% file: math_commands.tex
\usepackage{amsmath,amsfonts,bm}

\def\eqref#1{equation~\ref{#1}}

\def\1{\bm{1}}

\DeclareMathAlphabet{\mathsfit}{\encodingdefault}{\sfdefault}{m}{sl}
\SetMathAlphabet{\mathsfit}{bold}{\encodingdefault}{\sfdefault}{bx}{n}

